\documentclass[10pt,twocolumn,letterpaper]{article}

\usepackage{cvpr}              

\usepackage{minitoc}

\usepackage{pifont}

\usepackage{array} 
\newcolumntype{C}[1]{>{\centering\arraybackslash}p{#1}} 

\usepackage[table]{xcolor}
\definecolor{rowlight}{RGB}{240,244,255} 
\definecolor{ourbg}{RGB}{245,240,255}

\usepackage{url}

\usepackage{marvosym} 

\makeatletter
\newcommand\emailfootnotetext[1]{%
  \begingroup
    \renewcommand\thefootnote{}
    \footnotetext{\Letter~\href{mailto:#1}{#1}}%
    \addtocounter{footnote}{-1}
  \endgroup
}

\newcommand\emailthanks[1]{%
  \begingroup
    \renewcommand\thefootnote{}
    \footnotemark
  \endgroup
  \protected@xdef\@thanks{\@thanks\protect\emailfootnotetext{#1}}%
}
\makeatother

\definecolor{cvprblue}{rgb}{0.21,0.49,0.74}
\usepackage[pagebackref,breaklinks,colorlinks,allcolors=cvprblue]{hyperref}

\usepackage[dvipsnames]{xcolor} 

\usepackage{lipsum}

\usepackage{dblfloatfix}
\usepackage{multicol}
\usepackage{multirow}
\usepackage{xcolor}
\usepackage{makecell}
\usepackage{tabularx}
\usepackage{array}
\usepackage{booktabs}
\usepackage{colortbl}
\usepackage{graphicx}
\usepackage{adjustbox}

\usepackage[most]{tcolorbox} 
\usepackage{caption}
\newcommand{\ourmodel}{Video-CoM}
\newcommand{\ourdata}{Video-CoM-Instruct}
\newcommand{\ourbench}{Video-CoM-Bench}
\newcolumntype{C}[1]{>{\centering\arraybackslash}m{#1}}

\newcommand{\HeadH}{1.8cm} 
\newcommand{\rotheads}[1]{%
  \adjustbox{valign=b}{\rotatebox[origin=bl]{90}{\strut #1}}%
  \rule{0pt}{\HeadH}%
}
\newlength{\Wmodel}
\newlength{\Whigh}

\newcolumntype{L}[1]{>{\raggedright\arraybackslash}p{#1}}
\newcolumntype{C}[1]{>{\centering\arraybackslash}p{#1}}
\newcolumntype{Y}{>{\centering\arraybackslash}X}

\newcolumntype{A}[1]{>{\raggedright\arraybackslash}p{#1}} 
\newcolumntype{B}{>{\centering\arraybackslash}X}  
\usepackage{pifont} 
\newcommand{\ccmark}{\ding{51}} 
\newcommand{\xxmark}{\ding{55}} 

\def\paperID{} 
\def\confName{CVPR}
\def\confYear{2026}

\title{Video-CoM: Interactive Video Reasoning via Chain of Manipulations}

\author{
  Hanoona Rasheed$^{1}$\textsuperscript{\Letter}\emailthanks{hanoona.bangalath@mbzuai.ac.ae},~~
  Mohammed Zumri$^{1}$,~~
  Muhammad Maaz$^{1}$ \\
  Ming-Hsuan Yang$^{2,3}$,~~
  Fahad Shahbaz Khan$^{1,4}$,~~
  Salman Khan$^{1,5}$\\[0.25cm]
  \fontsize{10.5pt}{12pt}\selectfont
  $^{1}$Mohamed bin Zayed University of AI, $^{2}$University of California Merced\\
  \fontsize{10.5pt}{12pt}\selectfont
  $^{3}$Google Research, $^{4}$Linköping University, $^{5}$Australian National University\\
  {\hypersetup{}
    \fontsize{11.5pt}{12pt}\selectfont
    \href{https://github.com/mbzuai-oryx/Video-CoM}{https://github.com/mbzuai-oryx/Video-CoM}
  }
}

\begin{document}
\doparttoc 
\faketableofcontents 

\maketitle
\begin{abstract}
Recent multimodal large language models (MLLMs) have advanced video understanding, yet most still `\textit{think about videos}' i.e., once a video is encoded, reasoning unfolds entirely in text, treating visual input as a static context. 
This passive paradigm creates a \textit{semantic bottleneck}: models cannot rewatch, refocus, or verify evidence, leading to shallow visual reasoning on tasks requiring fine-grained spatio-temporal understanding. 
In this work, we introduce \textit{Interactive Video Reasoning}, a new paradigm that transforms video into an active cognitive workspace, enabling models to `\textit{think with videos}'. 
Our model, \textit{\ourmodel}, reasons through a \textit{Chain of Manipulations (CoM)}, performing iterative visual actions to gather and refine evidence. 
To support this behavior, we construct \textit{\ourdata}, an 18K instruction-tuning dataset curated for multi-step manipulation reasoning. 
Beyond supervised learning, we further optimize the manipulation policy via reinforcement learning with reasoning-aware Group Relative Policy Optimization (GRPO). 
Unlike prior work that relies solely on sparse answer rewards, our method introduces step-level reasoning rewards, guiding the model toward grounded and consistent reasoning. 
\ourmodel~achieves strong results across nine video reasoning benchmarks, improving average performance by 3.6\% over recent state-of-the-art models, while training on only 25K SFT and 3K GRPO video samples, significantly fewer than comparable large-scale models.
Ablation studies demonstrate that reasoning-aware rewards improve both accuracy and interpretability. 
\end{abstract}
    
\section{Introduction}
\label{sec:intro}

\begin{figure*}[t!]
  \centering
  \includegraphics[width=\textwidth]{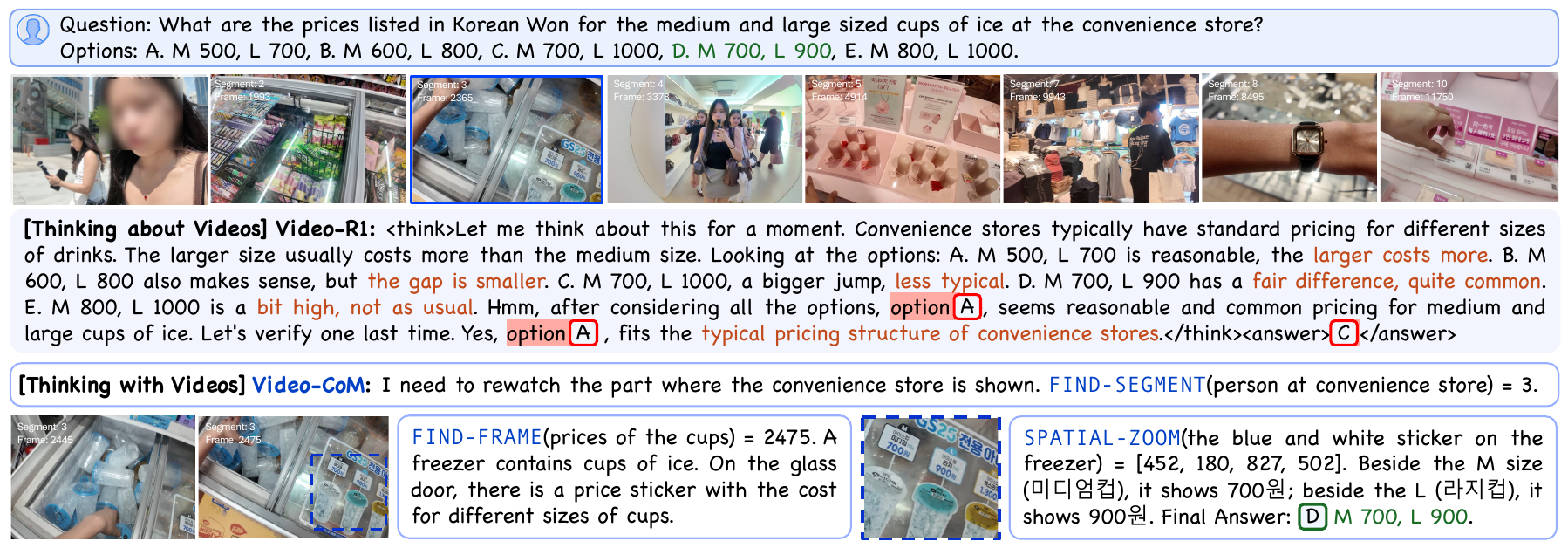}
  \vspace{-1.5em}
  \caption{
    Most existing video reasoning models \textit{think about videos} rather than \textit{think with them}.
    Once a video is encoded, reasoning unfolds purely in text, causing thinking tokens to drift toward world knowledge rather than visual evidence. Moreover, these models are typically trained with sparse accuracy rewards that supervise only the final answer, leading to inconsistencies between reasoning steps and predictions (top, \textit{Video-R1~\cite{feng2025videor1}}). We address these limitations with \textit{Interactive Video Reasoning}, where our model (bottom, \textit{\ourmodel}) reasons \textit{with videos} through a coherent \textit{chain of manipulations}, actively gathering and integrating visual evidence throughout reasoning.
    }
  \label{fig:motivation}
  \label{fig1_intro}
  \vspace{-1em}
\end{figure*}

Recent advances in multimodal large language models (MLLMs) have substantially improved video understanding, enabling models to describe, answer, and reason about visual content~\cite{zhu2025internvl3, bai2025qwen2.5VL, cho2025PerceptionLM, comanici2025gemini}. However, despite impressive scale and accuracy, most existing video reasoning models~\cite{feng2025videor1, li2025videochatr1, wang2025videorft, dang2025reinforcing} are driven by the paradigm of \textit{thinking about videos} rather than \emph{thinking with them}. Once a video is encoded, reasoning unfolds purely in text. The video acts as a static context, compressed into a single feature representation, while the subsequent reasoning unfolds only in the language space. This static treatment of perception creates a \textit{semantic bottleneck}: the model loses the ability to revisit, verify, or refine visual evidence. As a result, current models~\cite{feng2025videor1, li2025videochatr1} often fail on tasks requiring fine-grained temporal understanding, precise spatial reasoning, or multi-step evidence aggregation. Specifically, their thinking tokens tend to drift toward the world knowledge rather than the visual evidence in the video~(See Figure~\ref{fig:motivation} and Figure~\ref{fig:visual_attention}).

Humans, in contrast, \emph{think with videos} through an active perceptual loop. Rather than passively encoding visual input, we iteratively observe, pause, rewatch, and focus on different regions to gather and verify evidence. Figure~\ref{fig:motivation} illustrates this behavior. In the first example, answering, \textit{“what are the prices listed on the cups of ice?”} requires locating the brief moment at the convenience store, pausing on a frame where the prices are legible, and zooming into the small text on the freezer door. Models that reason primarily in text often miss such localized evidence, relying instead on general associations or prior knowledge, which leads to inconsistent and weakly grounded visual reasoning.

Motivated by this, we propose \textit{Interactive Video Reasoning}, a paradigm that transforms video from a passive input into a dynamic cognitive workspace. At the core of our approach lies a \textit{chain of manipulations} (CoM) mechanism that enables our model, \textit{\textbf{\ourmodel}}, to actively interact with the video throughout the reasoning process. Instead of producing an answer directly from a static encoding, the model performs a sequence of informed visual actions. This notion of manipulation-based reasoning extends ideas from thinking with images~\cite{openai2025thinkingwithimages} and chain of image manipulations~\cite{qi2024cogcom} to the temporal domain. While prior works~\cite{openai2025thinkingwithimages, zheng2025deepeyes, gupta2023visual, suris2023vipergpt, shen2024zoomeye, su2025thinkingimagesmultimodalreasoning} have shown that visual reasoning improves when models can perform active perception, cropping or segmenting images during reasoning, our setting introduces new challenges. Videos unfold over time, and reasoning often depends on temporally localized events, short-lived details, or subtle motions. To mimic how humans extract such information, we introduce \textit{three} atomic manipulations, \textit{i}) \textit{find-segment} to revisit short temporal windows, \textit{ii}) \textit{find-frame}  to isolate key frames for closer inspection, and \textit{iii}) \textit{spatial-zoom} to examine fine details  (e.g., small text or objects). By composing these manipulations, the model constructs an interpretable reasoning trajectory, where each step produces concrete visual evidence.

Building such a capability requires data for complex video reasoning that requires active manipulations. Existing video-understanding datasets~\cite{feng2025videor1,caba2015activitynet,zhou2017youcookii, chen2024sharegpt4video,maaz2024video,xiao2021next,maaz2024videogpt+,zhang2024video} rarely demand this level of engagement. To drive manipulation-based video reasoning, we construct \textit{\ourdata}, an 18K high-quality instruction-tuning dataset designed to elicit the use of manipulations for multi-step reasoning. 
Each sample is carefully curated to require one or more visual operations, mirroring how humans extract localized evidence. The dataset teaches how to perform manipulations and exposes the model to a diverse set of reasoning trajectories, when to use each manipulation and how to chain them effectively across video contexts.

While high-quality data provides the foundation, supervised fine-tuning alone is insufficient. The manipulation trajectory is inherently task-dependent. The model must also learn how to plan and execute the right sequence of manipulations, conditioned on both the question and the video content. To achieve this, we adopt reinforcement learning with group relative policy optimization (GRPO)~\cite{shao2024deepseekmath} to directly optimize the manipulation policy. Existing video reasoning models~\cite{feng2025videor1, li2025videochatr1, wang2025videorft} are typically optimized with sparse accuracy rewards that supervise only the final answer, providing no feedback on the intermediate reasoning process. As shown in Figure~\ref{fig:motivation}, such text-based reasoning often lacks grounding in visual evidence, and the final answer tokens are sometimes inconsistent with the preceding reasoning.

We address this limitation through a \textit{step-level reasoning reward} that evaluates intermediate actions i.e., temporal IoU for find-segment, frame recall for find-frame, and spatial IoU for spatial-zoom. This step-level supervision credits partial progress even when the final answer is incorrect, encouraging grounded and consistent reasoning. Empirically, \ourmodel~demonstrates consistent improvements on nine diverse video benchmarks, including generic, long-form understanding as well as challenging reasoning datasets. Trained on only 25K SFT and 3K GRPO video samples, it matches or exceeds models trained on far larger datasets~\cite{feng2025videor1, yan2025videochat_r1.5}. Our ablations further confirm that reasoning rewards enhance both accuracy and coherence, 
enabling the model to think with videos, not just about them. 

The main contributions of this work are:
\begin{itemize}[leftmargin=1.5em, itemsep=0pt, topsep=2pt]
    \item We present \textit{\ourmodel}, a multimodal reasoning model for \textit{interactive video reasoning} that actively gathers visual evidence through a chain of manipulations (\S\ref{sub_sec:com}).
    \item We curate \textit{\ourdata}, an 18K manipulation-driven instruction-tuning dataset that captures diverse reasoning trajectories for interactive reasoning (\S\ref{sec:data}).
    \item We propose a \textit{reasoning-aware GRPO} that directly optimizes the manipulation policy using \textit{reasoning rewards} that shape optimal manipulation trajectories (\S\ref{sub_sec:reasoning_reward}).
    \item Despite using substantially less training data than prior large-scale approaches, our approach achieves competitive performance across nine video benchmarks (\S\ref{sub_sec:results}).
\end{itemize}
\vspace{-0.em}
\section{Related Work}
\label{sec:related_work}

\noindent \textbf{Static Visual Reasoning.}
Early studies on chain-of-thought (CoT) reasoning~\cite{10.5555/3600270.3602070,10.5555/3600270.3601883} showed that prompting LLMs to reason step by step improves performance on multi-hop reasoning tasks. This concept was later extended to the visual domain through visual CoT methods~\cite{liu2023visual,zhang2023multimodal,xu2025llava}, enabling MLLMs to perform structured grounded reasoning. However, perception in these frameworks remains static; visual features are encoded once and then consumed as fixed context during reasoning. As noted by~\cite{su2025thinkingimagesmultimodalreasoning}, this limits the ability to iteratively verify or refine visual understanding. Recent works move beyond this paradigm, from models that \textit{think about images} to those that \textit{think with them}, where perception actively interacts with reasoning~\cite{Zeng2022SocraticMC,yang2023mmreact,wu2023visual,liu2024chain,shen2024zoomeye,DBLP:journals/corr/abs-2505-18842,gupta2023visual,suris2023vipergpt,hu2024visual,fu2025refocus,qi2024cogcom,zhao2025cot,li2025imagine,xu2025visual, shao2024visual}. However, extending CoT frameworks to videos introduces additional challenges in modeling temporal dependencies, dynamic events, and fine-grained cues across frames.

\noindent \textbf{Interactive Video Reasoning.}
Most video reasoning models~\cite{feng2025videor1,li2025videochatr1,wang2025videorft,dang2025reinforcing} still treat videos as static context, performing reasoning entirely in the language space once the video is encoded. VideoChat-R1.5~\cite{yan2025videochat_r1.5} employs iterative perception with test-time scaling, repeatedly refining its spatio-temporal focus. In contrast, interactive video reasoning methods enable dynamic spatio-temporal interaction, allowing models to gather new visual evidence during reasoning. More recent approaches integrate this, where models actively retrieve visual information through tool calls as reasoning unfolds. VITAL~\cite{zhang2025thinkingvideosmultimodaltoolaugmented} samples new frames on demand and incorporates them into the CoT. VideoExplorer~\cite{yuan2025thinkvideosagenticlongvideo} adopts a similar strategy, grounding relevant video segments and integrating retrieved frames at each reasoning step. FrameMind~\cite{ge2025framemind} proposes interleaving reasoning and frames, allowing the model to pause and identify missing visual evidence. These approaches mark an important step toward interactive reasoning, but they generally rely solely on temporal resampling or frame retrieval, without incorporating any spatial operations, which restricts their ability to reason jointly over time and space.

Unlike prior interactive video reasoning approaches, which perform repetitive frame retrieval loops within a segment, our Chain of Manipulations (CoM) defines a structured and compositional reasoning process with temporal and spatial grounding. Each manipulation in our case represents a distinct atomic visual action and chaining them forms an interpretable trajectory that explicitly grounds each reasoning step in localized evidence. In contrast, prior methods typically employ a temporal manipulation type, either temporal re-sampling or frame isolation, without demonstrating how both temporal and spatial manipulations can be composed to perform multi-level reasoning. Furthermore, our approach not only composes these manipulations but also learns to refine reasoning trajectories through reasoning-aware reinforcement learning (RA-GRPO), thereby enabling consistent visual grounding and improved coherence across reasoning steps, an aspect not addressed in prior “thinking with videos” frameworks.

\section{Method}
\label{sec:method}

\begin{figure*}[t!]
  \centering
  \includegraphics[width=\textwidth]{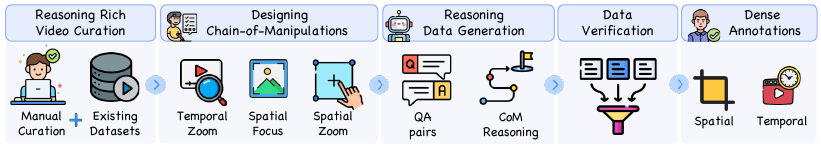}
  \vspace{-1.5em}
  \caption{\textbf{Overview of the \ourdata-18K dataset pipeline.} The dataset is built through five stages: \textit{i)} Curating videos for active reasoning, combining information-dense videos from existing datasets and manually curated clips; \textit{ii)} Designing chain-of-manipulations, defining the atomic operations (find-segment, find-frame, spatial-zoom); \textit{iii)} Interactive video reasoning data generation, where Gemini-2.5-Pro produces manipulation-targeted QA pairs; \textit{iv)} Data verification using model agreement filtering ensuring quality; and \textit{v)} Dense spatio-temporal annotation, which enables step-level reasoning rewards for GRPO. The resulting \textit{\ourdata} dataset contains 18K QA pairs (15K for SFT and 3K for GRPO), supporting interactive, manipulation-based video reasoning.}
  \label{fig:dataset_pipeline}
  \vspace{-1em}
\end{figure*}

Our goal is to enable multimodal large language models (MLLMs) to \textit{think with videos} rather than merely \textit{think about them}. Existing video reasoning models typically encode a video once to obtain a static representation, after which reasoning unfolds entirely in text. This static formulation prevents the model from revisiting, refining, or grounding its reasoning in visual evidence. To overcome this limitation, we propose \textit{Interactive Video Reasoning}, which carefully integrates \textit{three} components that address the key requirements of this capability:
\textbf{(1)} a \textit{Chain of Manipulations} (CoM) mechanism (\S\ref{sub_sec:com}) that allows the model to actively interact with the video through a sequence of visual actions; 
\textbf{(2)} a curated \textit{Dataset for Interactive Video Reasoning} (\S\ref{sec:data}) that provides the  necessary supervision to elicit such behavior through manipulation-driven instruction tuning; and
\textbf{(3)} \textit{Reasoning-Aware Group Relative Policy Optimization} objective (\S\ref{sub_sec:reasoning_reward}) that optimizes the model to plan and execute effective manipulation trajectories through step-level reasoning rewards.

\subsection{Modeling the Chain of Manipulations}
\label{sub_sec:com}

The core idea of our framework is to model reasoning as a multi-turn interaction, where each step corresponds to an informed visual manipulation. At every turn, the model selects one of three atomic manipulations: \textit{find-segment}, \textit{find-frame}, or \textit{spatial-zoom}, to actively gather localized visual evidence. At reasoning step $i$, given the textual context $T_i$ and current visual input $\mathcal{V}_i$, the model generates an output,
$$\mathcal{O}_i = f_{\text{MLLM}}(T_i, \mathcal{V}_i),$$
where $\mathcal{O}_i$ includes both reasoning tokens and a predicted manipulation $m_i$. Each manipulation is then executed by $g$ to perform the corresponding visual action and the resulting observation is fed back to the model for continued reasoning, $\mathcal{V}_{i+1} = g(\mathcal{V}_i, m_i)$. This process forms an iterative loop that continues until the model produces a final answer or reaches a predefined limit of $N$ reasoning rounds, resulting in a reasoning trajectory, $$\tau = \{(T_1, \mathcal{V}_1, m_1), (T_2, \mathcal{V}_2, m_2), \ldots, (T_N, \mathcal{V}_N, m_N)\}$$

Each video is pre-segmented into equal-length intervals, with both segment and frame indices overlaid directly onto the frames. This design makes temporal references explicit and allows the model to learn manipulation-based reasoning naturally, without relying on explicit timestamp prediction. Each manipulation represents a distinct mode of visual interaction that simulates how humans actively inspect videos. 
\textit{\textbf{(i)}} \textit{Find-segment} $g_{\text{clip}}$ retrieves a short temporal clip from the original video $\mathcal{V}_0$ given a segment index $s$, as 
$\mathcal{V}_{i+1} = g_{\text{clip}}(\mathcal{V}_0, s)$. 
It enables the model to revisit and rewatch rapid actions or critical moments essential for understanding event order or temporal transitions. 
\textit{\textbf{(ii)}} \textit{Find-frame} $g_{\text{frame}}$ isolates a single high-resolution frame from $\mathcal{V}_0$ given a frame index $f$, expressed as 
$\mathcal{V}_{i+1} = g_{\text{frame}}(\mathcal{V}_0, f)$. 
This allows the model to pause on a specific moment for spatial reasoning tasks such as object presence, counting, or reading small text. 
\textit{\textbf{(iii)}} \textit{Spatial-zoom} $g_{\text{zoom}}$ operates on the current frame, using a frame index $f$ and a bounding box $b$, expressed as 
$\mathcal{V}_{i+1} = g_{\text{zoom}}(\mathcal{V}_i, f, b)$. 
It enables the model to crop or magnify a localized region to examine fine-grained details such as license plates, prices, or distant objects. Together, these manipulations emulate how humans pause, rewatch, and focus to extract and verify evidence, ensuring that each reasoning step remains grounded in concrete visual observations rather than unfolding purely in text.

\begin{figure*}[t]
  \centering
  \includegraphics[width=\textwidth]{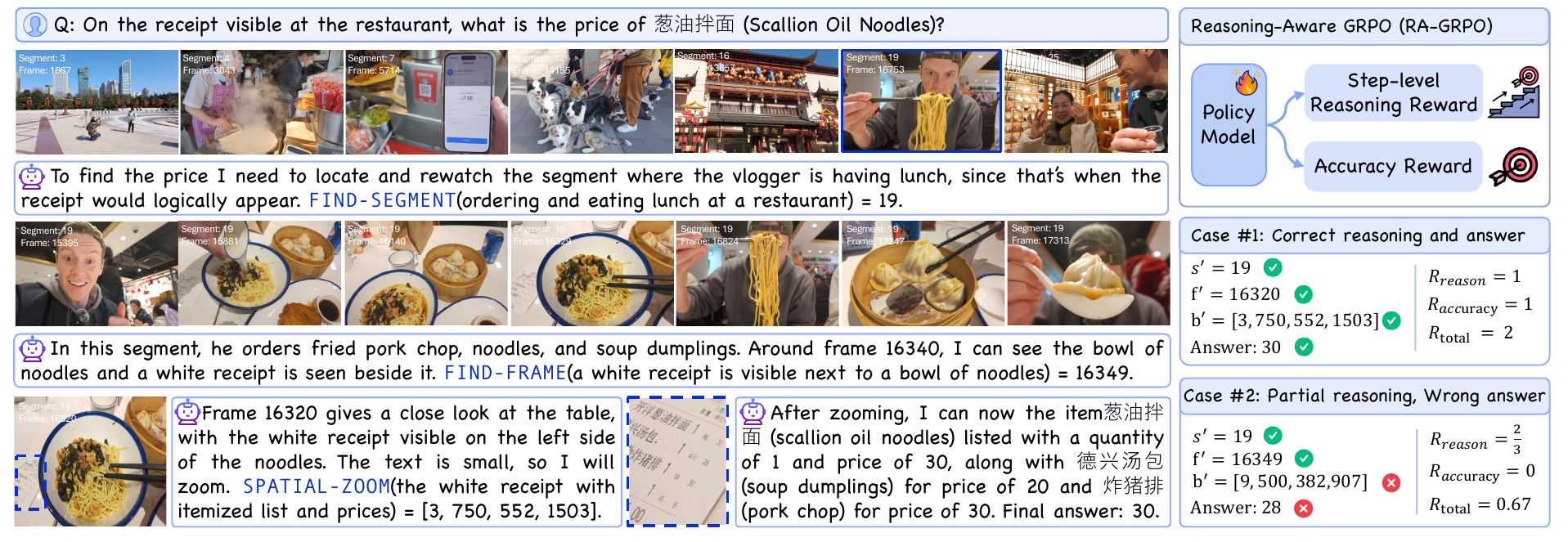}
  \vspace{-1.5em}
  \caption{
  \textbf{Overview of Interactive Video Reasoning and the proposed Reasoning-Aware GRPO (RA-GRPO) framework.} 
  The model, \textit{\ourmodel}, reasons \textit{with videos} through a \textit{Chain of Manipulations} (CoM), iteratively applying visual operations to gather and refine visual evidence. 
  Each manipulation produces new observations that are fed back into the model for continued reasoning. 
  Because multiple manipulation trajectories can lead to the same correct answer, accuracy-only rewards cannot distinguish grounded from spurious reasoning paths. 
  RA-GRPO resolves this by introducing \textit{step-level reasoning rewards} that explicitly evaluate each manipulation. 
  The examples illustrate how intermediate reasoning steps are verified, providing partial credit for correct intermediate actions and guiding the model toward consistent, visually grounded reasoning trajectories.
  }
  \label{fig:method_fig}
  \vspace{-1em}
\end{figure*}

\subsection{Dataset for Interactive Video Reasoning}
\label{sec:data}

Training a model to reason through manipulations requires data that explicitly elicits such behavior. Existing instruction-tuning datasets~\cite{zhang2024video, chen2024sharegpt4video, maaz2024video, maaz2024videogpt+} fall short in two key aspects. First, their question-answer (QA) pairs focus primarily on global video understanding, without encouraging localized evidence gathering across time and space. Second, the underlying video sources are often contextually uniform, typically depicting a single continuous scene~\cite{caba2015activitynet, zhou2017youcookii, xiao2021next, yi2019clevrer}, limiting temporal and visual diversity. In our setting, driving manipulation-based video reasoning requires QAs that target complex, multi-step reasoning involving active manipulations. However, constructing such QAs in turn demands information-dense videos with diverse scenes, fine-grained details, and temporally localized events. We therefore address both aspects through curated video selection and manipulation-driven QA construction, which together form the foundation of our 
dataset pipeline, as illustrated in Figure~\ref{fig:dataset_pipeline}.

\noindent \textbf{Curating Videos for Active Reasoning.}
Our dataset builds on two complementary video sources: large-scale existing datasets for SFT and a curated collection of high-complexity videos for RL. For SFT, we sample videos from Video-LLaVA-178K~\cite{zhang2024video}, which aggregates a diverse set of videos from multiple datasets~\cite{caba2015activitynet, zhou2017youcookii, sigurdsson2016hollywood_charades, xue2022advancing-hdvila, wang2023internvid, shang2019annotating, zhu2023languagebind, kay2017kinetics, grauman2022ego4d}. To identify videos suitable for manipulation-driven QA generation, we estimate their information density, which reflects the richness of visual and temporal details available for localized evidence gathering. Video captions are first generated using Qwen2.5-VL-72B~\cite{bai2025qwen2.5VL}, and Qwen3-235B-A22B~\cite{yang2025qwen3} assigns an information-density score. We then select approximately 9K videos with the highest scores, ensuring diversity across data sources. Details on video distribution and the scoring prompts are provided in Appendix~\ref{appendix_sec:data}.

For RL, we carefully curate 70 high-quality, multi-context videos from YouTube that exhibit rich temporal and spatial complexity. We target clips featuring: (\textit{i}) rapid or sequential actions requiring precise temporal localization and rewatching, (\textit{ii}) small or transient visual elements that prompt frame-level reasoning, such as counting or reading text, and (\textit{iii}) fine-grained details that benefit from spatial zooming, such as reading signs, license plates, or prices. These characteristics make them particularly effective for constructing manipulation-driven questions, where the model must decide when to revisit a segment, which frame to inspect, and where to zoom. In total, our dataset comprises approximately 9K videos, drawn from existing datasets and our manually curated collection.

\noindent \textbf{Designing Chain-of-Manipulations.} 
To identify common reasoning trajectories required for manipulation-based video reasoning, we manually study a set of representative questions that exhibit multi-step reasoning. This analysis reveals recurring reasoning patterns and the specific operations needed to gather visual evidence effectively. Based on these findings, we define three atomic manipulations: \textit{find-segment}, \textit{find-frame}, and \textit{spatial-zoom}, and identify a set of common chains of manipulations (CoM), such as \textit{find-frame}→\textit{spatial-zoom} for reading small text (for e.g., prices on a label), as shown in Figure~\ref{fig:motivation} or \textit{find-segment} for counting brief, repeated actions within a longer sequence. We then design a systematic in-context learning prompt that describes the purpose, inputs, outputs, and typical use cases of each manipulation, along with reasoning trajectories and how intermediate states evolve as new evidence is gathered.

\noindent \textbf{Interactive Video Reasoning Data Generation.} With the curated videos and defined chains of manipulations, we next construct a high-quality dataset that elicits active video reasoning. Starting from 9K videos, we use Gemini-2.5-Pro~\cite{comanici2025gemini} to generate 3-5 QA pairs per video, producing 25K QA pairs in total. Each question is designed to target a specific CoM. To ensure quality, we re-evaluate every generated question in two formats, multiple-choice and open-ended, which provides an effective mechanism for filtering ambiguous or unanswerable samples. After this verification, roughly 75\% of the data are retained, resulting in 18K high-quality QA pairs that include the question, answer, MCQ options, and CoM reasoning trace. For spatial grounding, we find that predicting bounding boxes directly from the full video is unreliable. We therefore re-annotate all spatial localizations using the corresponding frame as input and keep only those with high agreement between Gemini and InternVL3-78B~\cite{zhu2025internvl3}. The prompts used for QA generation and verification are provided in Appendix~\ref{appendix_sec:data} Table~\ref{tab:data_qa_generation_prompt} and~\ref{tab:data_qa_verification_prompt}.

\noindent \textbf{\ourdata.} Building on the above pipeline, we construct \textit{\ourdata}, a high-quality collection of 18K QA pairs designed to enable interactive, manipulation-based video reasoning. Of these, 16.5K are generated from videos sampled from existing datasets and 1.5K from the manually curated collection. We use 15K QA pairs for SFT and set aside 3K for GRPO. All samples from the curated videos are included in the GRPO subset, along with an additional 1.5K selected from existing sources. To sample these, we generate eight predictions per question using Qwen2.5-VL-72B~\cite{bai2025qwen2.5VL} and retain only examples that are neither trivially solvable nor consistently failed across all predictions.

\noindent \textbf{Dense Annotation Enabling Step-Level Reasoning Rewards.} To enable reasoning-aware optimization through GRPO, we enrich the 3K GRPO samples with dense temporal and spatial annotations. Each manipulation step in the target CoM is comprehensively labeled, allowing rewards to be assigned at the step level rather than only for final answers. For temporal reasoning, we annotate all valid segments and frames for each \textit{find-segment} and \textit{find-frame} operation, since the same piece of evidence may appear across multiple moments in a video. For spatial reasoning, we manually annotate bounding boxes on representative frames and track them using SAM2.1~\cite{ravi2024sam} to form tubelets that capture the full temporal span where the target object appears. These dense annotations ensure consistent grounding of visual evidence across time, enabling the use of step-level rewards for interactive, manipulation-based reasoning.

\noindent \textbf{Benchmark for Manipulation-Based Video Reasoning.} While existing benchmarks such as Video-MMMU~\cite{hu2025video_mmmu} and ScienceVid~\cite{deng2025scivideobench} emphasize compositional or scientific reasoning, they do not explicitly assess the ability to perform \textit{manipulation-based reasoning}. To rigorously evaluate the effectiveness of interactive video reasoning, we construct \textit{\ourbench}, a dedicated benchmark targeting manipulation-centric reasoning. The data generation process follows the same pipeline as the  {\textit{\ourdata}} GRPO subset. Human annotators verify each QA pair, multiple-choice options, reasoning trace and dense spatio-temporal localization to ensure annotation quality. In total, \textit{\ourbench} comprises 1K high-quality samples for evaluating manipulation-based video reasoning.

\subsection{Reasoning-Aware Group Relative Policy Optimization (RA-GRPO)}
\label{sub_sec:reasoning_reward}

Most existing video reasoning models are optimized through outcome-based supervision, where rewards are assigned solely to the final answer. While this objective enforces correctness at the end of reasoning, it provides no guidance for the intermediate reasoning steps. As a result, the model’s reasoning trajectory often drifts away from the visual evidence, relying on world knowledge rather than the video, and may generate reasoning that are inconsistent with the final prediction~(see Figure~\ref{fig:motivation} and Figure~\ref{fig:visual_attention}). In contrast, our framework performs \textit{chain-of-manipulations}-based reasoning, where each step such as finding segments, isolating frames, or zooming spatially produces new and verifiable visual evidence. Since multiple valid manipulation sequences can yield the same correct answer, accuracy-only rewards cannot distinguish between grounded and spurious reasoning paths.

To address this limitation, we introduce \textit{Reasoning-Aware Group Relative Policy Optimization} (RA-GRPO), an extension of GRPO that incorporates \textit{step-level reasoning rewards} to explicitly evaluate the correctness of each visual manipulation. For every video-question pair, we generate a group of reasoning trajectories sampled from the current policy. Each trajectory consists of a complete reasoning sequence, where at each step the model produces an output $\mathcal{O}_i$ containing both textual reasoning and a predicted manipulation $m_i$. Chaining the intermediate outputs yields a chain of manipulations $CoM = \{m_1, m_2, \ldots, m_{N-1}\}$, while the final output $\mathcal{O}_N$ concludes the reasoning process, from which the answer $\hat{y}$ is extracted from its textual content.

For each question, supervision is restricted to the manipulation types relevant to its annotated reasoning trajectory. RA-GRPO evaluates correspondence within this subset when computing the reasoning reward. Specifically, we define the \textbf{\textit{reasoning reward}} $R_{\text{reason}}$, which measures the proportion of correct manipulations within a reasoning trajectory. For each predicted manipulation $m_i'$ among $N$ predicted manipulations, correctness is defined as
\begin{equation}
c_i =
\begin{cases}
1, & 
\begin{cases}
s_i' \in \mathcal{S}^*, & \text{for } \textit{find-segment}, \\
f_i' \in \mathcal{F}^*, & \text{for } \textit{find-frame}, \\
\text{IoU}(b_i', b^*) \ge \tau_b, & \text{for } \textit{spatial-zoom}, 
\end{cases} \\
0, & \text{otherwise.}
\end{cases}
\label{eq:ci}
\end{equation}
\noindent
where $\mathcal{S}^*$, $\mathcal{F}^*$, and $\mathcal{B}^*$ denote the sets of annotated valid segments, frame indices, and bounding boxes, respectively, and $\tau_b$ is the IoU threshold with $b^* \in \mathcal{B}^*$. The reasoning reward is then computed as the mean correctness over all predicted manipulations:
\begin{equation}
R_{\text{reason}} = \frac{1}{N} \sum_{i=1}^{N} c_i.
\label{eq:r_reason}
\end{equation}

This step-level design enables dense supervision, rewarding reasoning trajectories that perform correct manipulations even when the final answer is wrong.
In addition, we define an \textbf{accuracy reward} $R_{\text{accuracy}}$ that evaluates the final answer $\hat{y}$ correctness against the ground truth $y^*$:
\begin{equation}
R_{\text{accuracy}} =
\begin{cases}
1, & \text{if } \hat{y} = y^*, \\
0, & \text{otherwise.}
\end{cases}
\label{eq:r_ans}
\end{equation}
\noindent
\begin{equation}
R = R_{\text{accuracy}} + R_{\text{reason}}.
\label{eq:R}
\end{equation}
\noindent Together, these terms define the total reward for each trajectory (See Figure~\ref{fig:method_fig}). This design enables RA-GRPO to assign partial credit to trajectories that follow correct reasoning even when the final answer is incorrect, while still emphasizing trajectories that are both accurate and visually grounded. By shaping the reward in this way, RA-GRPO encourages the model to develop reasoning policies that integrate factual correctness with consistent visual grounding.
\begin{table*}[!t]
\centering
\begingroup
\setlength{\extrarowheight}{1pt}
\begin{adjustbox}{max width=\textwidth}
\begin{tabularx}{\textwidth}{
  L{\Wmodel}|
  >{\columncolor{blue!6}}C{\Whigh}|
  *{6}{Y}|
  >{\columncolor{blue!6}}C{\Whigh}|
  *{3}{Y}|
  >{\columncolor{blue!6}}C{\Whigh}
}
\toprule
& \multicolumn{8}{c|}{\textbf{Reasoning Benchmarks}} & \multicolumn{4}{c}{\textbf{Generic Benchmarks}} \\
\cmidrule(lr){3-9}\cmidrule(lr){10-13}
\textbf{Model} &
\rotheads{\textbf{VCoM-Bench}} &
\rotheads{MMMU/Perc} &
\rotheads{MMMU/Avg} &
\rotheads{ScienceVid} &
\rotheads{VideoMath} &
\rotheads{MMVU-Val} &
\rotheads{Minerva} &
\rotheads{\textbf{Avg-Reason}} &
\rotheads{VideoMME} &
\rotheads{TempComp} &
\rotheads{MLVU} &
\rotheads{\textbf{Avg-Generic}} \\
\midrule
Video-R1~\cite{feng2025videor1}             & 57.1 & 62.0 & 47.2 & \underline{26.1} & 23.3 & 63.8 & 29.1 & 37.9 & \underline{59.3} & 66.9 & \textbf{60.9} & 62.4 \\
VideoChat-R1~\cite{li2025videochatr1}       & 56.3 & 67.0 & \textbf{50.3} & 25.5 & 23.3 & 64.8 & 30.3 & 38.8 & 58.2 & \textbf{72.5} & 57.0 & 62.6 \\
VideoChat-R1.5~\cite{yan2025videochat_r1.5} & 59.9 & 65.6 & 48.3 & 23.8 & \underline{25.7} & \textbf{67.8} & \underline{31.0} & \underline{39.3} & 59.1 & \underline{72.3} & 58.2 & 63.2 \\
VideoRFT~\cite{wang2025videorft}            & 60.2 & \underline{68.0} & 48.9 & 23.8 & 25.2 & \underline{66.7} & 29.2 & 38.8 & 60.1 & 70.4 & \underline{59.7} & \underline{63.4} \\
\midrule
\rowcolor{blue!6} Video-CoM                 & \textbf{68.7} & \textbf{70.0} & \underline{50.2} & \textbf{27.6} & \textbf{27.8} & 65.4 & \textbf{31.7} & \textbf{40.5} & \textbf{59.4} & 71.3 & \textbf{60.9} & \textbf{63.9} \\
\bottomrule
\end{tabularx}
\end{adjustbox}
\endgroup
\caption{\textbf{Comparison across nine video understanding benchmarks.} \textit{\ourmodel} achieves the highest average score on five reasoning benchmarks and remains competitive on generic ones. The largest gain is observed on {\ourbench}, which requires multi-step evidence gathering, highlighting that \textit{interactive video reasoning} through chain-of-manipulations leads to stronger and grounded reasoning.}
\label{tab:main_results}
\end{table*}
\vspace{-0.7em}
\section{Experiments}
\label{sec:experiments}

\subsection{Training Details}
\label{sec:training_details}
We adopt \textit{Qwen2.5-VL-7B-Instruct}~\cite{bai2025qwen2.5VL} as the base MLLM and train \textit{\ourmodel} in two stages. \noindent \textbf{Stage I (SFT).} We perform SFT on \textit{\ourdata}, which provides 15K high-quality samples for chain-of-manipulations reasoning. To enhance temporal understanding, we add 9K ActivityNet~\cite{caba2015activitynet} samples trained without manipulations, offers dense temporal supervision for improving frame localization. We further include 180K Visual-CoT~\cite{shao2024visual} samples focusing on image-based QA with spatial grounding, which strengthen the model’s \textit{spatial-zoom} capability. To balance optimization across modalities, Visual-CoT data are trained with proportionally larger batch sizes, ensuring comparable update steps between image and video data. The final SFT dataset comprises 25K video and 180K image samples.

\noindent \textbf{Stage II (RA-GRPO).} In the second stage, we apply \textit{reasoning-aware GRPO} (RA-GRPO) on the 3K GRPO subset of \textit{\ourdata}. Each sample includes dense temporal and spatial annotations for every manipulation step, allowing computation of both accuracy and step-level reasoning rewards. RA-GRPO encourages the model to reason with the video in an interactive manner through a coherent chain of manipulations, progressively gathering and integrating visual evidence at each step.

\subsection{Implementation Details}
\label{sec:implementation_details}
During training, videos are uniformly sampled at 2 FPS, and up to 32 frames are selected per video. Each frame is processed at a resolution of 360×420 pixels, and frame as well as segment indices are overlaid on the visual inputs to facilitate temporal localization. For \textit{find-segment}, up to 8 frames are uniformly sampled from the selected segment based on its duration. The number of manipulation turns is capped at 5 during both training and evaluation. All evaluations are conducted using the \texttt{lmms-eval}~\cite{zhang2025lmms} framework, following the same 2 FPS, 32 frame setting and employing the official prompts specified by each benchmark. 

We evaluate \textit{\ourmodel} across \textit{nine} video understanding and QA benchmarks, including five video-reasoning benchmarks, namely, VideoMMMU~\cite{hu2025video_mmmu}, MMVU-Val~\cite{mmvu2024}, Minerva~\cite{minerva2024}, ScienceVideoBench~\cite{deng2025scivideobench}, and VideoMathQA~\cite{rasheed2025videomathqa}, and three generic benchmarks (TempoCompass~\cite{tempcompass}, Video-MME~\cite{li2023videomme}, and MLVU~\cite{zhou2025mlvu}), where Video-MME spans short, medium, and long duration videos, and MLVU serves as a dedicated long-video benchmark. We further evaluate on our manipulation-specific benchmark, \textit{\ourbench}. We compare against four state-of-the-art reasoning models~\cite{feng2025videor1, li2025videochatr1, yan2025videochat_r1.5, wang2025videorft}, all built upon {Qwen2.5-VL-7B} and trained using two-stage SFT and GRPO pipelines. Direct comparison with concurrent interactive video-reasoning works~\cite{zhang2025thinkingvideosmultimodaltoolaugmented, yuan2025thinkvideosagenticlongvideo, ge2025framemind} is not included, as their models and code are not publicly available at submission time. Additional implementation details, hyperparameters and runtime comparison are provided in Appendix~\ref{appendix_sec:implementation}.

\subsection{Results}
\label{sub_sec:results}
\noindent \textbf{Main Results.}~Table~\ref{tab:main_results} summarizes the performance of \textit{\ourmodel} across nine video understanding and reasoning benchmarks. On the five \textbf{reasoning benchmarks}, \textit{\ourmodel} achieves the highest average score of $40.5$. VideoChat-R1.5 follows with $39.3$, while VideoRFT show similar trends. These results demonstrates that while RL optimization improves general reasoning, interactive, manipulation-based reasoning further enhances the model’s ability to identify and integrate relevant visual cues. On the three \textbf{generic video understanding benchmarks}, \textit{\ourmodel} remains competitive, reaching an average score of $63.9$, comparable to VideoRFT ($63.4$) and VideoChat-R1 ($63.2$). This shows that increasing interactivity with the video does not compromise broader video understanding capability. The most notable gain appears on the \textbf{manipulation-focused} \textit{\ourbench}, where \textit{\ourmodel} achieves $68.7$. This improvement highlights the model’s ability to reason \textit{with} the video iteratively, progressively integrating visual evidence. VideoRFT follows with $60.2$, benefiting from semantic alignment, while VideoChat-R1.5 reaches $59.9$ through multi-round perception. Notably, \textit{\ourmodel} attains these results using only $3$K RL samples, significantly fewer than the $18$K-$310$K used by comparable models~\cite{feng2025videor1, li2025videochatr1, yan2025videochat_r1.5, wang2025videorft}, highlighting the efficiency of our approach.
\begin{table}[t]
\centering
\begingroup
\setlength{\tabcolsep}{4pt}
\setlength{\extrarowheight}{2pt}
\renewcommand{\arraystretch}{1.05}
\begin{adjustbox}{width=\columnwidth,center}
\begin{tabular}{
  l|
  >{\columncolor{blue!6}}c|
  *{5}{c}|
  >{\columncolor{blue!6}}c
}
\toprule
\textbf{Model} &
\rotheads{\textbf{VCoM-Bench}} &
\rotheads{VidMMMU} &
\rotheads{VideoMath} &
\rotheads{ScienceVid} &
\rotheads{MMVU-Val} &
\rotheads{Minerva} &
\rotheads{\textbf{Avg-Reason}} \\
\midrule
SFT     & 64.0 & 47.3 & 26.7 & 23.8 & 59.4 & 30.1 & 37.5 \\
GRPO    & 66.7 & 47.3 & 27.3 & 25.9 & 62.7 & 31.2 & 38.9 \\
\rowcolor{blue!6} RA-GRPO & \textbf{68.7} & \textbf{50.2} & \textbf{27.6} & \textbf{27.8} & \textbf{65.4} & \textbf{31.7} & \textbf{40.5} \\
\bottomrule
\end{tabular}
\end{adjustbox}
\endgroup
\caption{\textbf{Ablation on reasoning-aware GRPO.}
Incorporating step-level reasoning rewards leads to consistent improvements across all benchmarks, indicating that refining intermediate reasoning steps enhances interactive video reasoning.}
\label{tab:ablations_ra_grpo}
\vspace{-0.5em}
\end{table}

\noindent \textbf{Effect of Reasoning-aware GRPO.} We evaluate the impact of the proposed \textit{reasoning-aware} GRPO by comparing three training stages: SFT, GRPO with accuracy-only rewards, and RA-GRPO with step-level reasoning rewards. As shown in Table~\ref{tab:ablations_ra_grpo}, SFT establishes the baseline for manipulation-based reasoning, achieving $64.0$ on \textit{\ourbench} and $37.5$ on average across the five reasoning benchmarks. Incorporating GRPO improves performance to $66.7$ and $38.9$. Introducing step-level reasoning rewards in RA-GRPO leads to a consistent improvement, reaching $68.7$ on \textit{\ourbench} and 40.5 on average. Accuracy-only rewards cannot distinguish between trajectories that are visually grounded and those that rely on spurious correlations. These results demonstrate how step-level evaluation addresses this limitation. RA-GRPO therefore promotes reasoning \textit{with the video} through a coherent chain of manipulations, progressively gathering and integrating visual evidence across reasoning steps.

\noindent\textbf{Effect of Visual Manipulations.} Table~\ref{tab:ablation_com}~evaluates the effect of each manipulation. We ablate it in \textit{\ourmodel} by withholding new visual inputs for that manipulation during training and inference. The text-only baseline performs reasoning without visual feedback. Introducing \textit{find-segment} shows the largest gain, highlighting the importance of temporal rewatching for fine-grained localization. Adding \textit{find-frame} enhances spatial reasoning by allowing the model to focus on key moments, while \textit{spatial-zoom} enables detailed inspection of small regions. Our analysis suggests that the impact of each manipulation correlates with its localization difficulty; temporal segments are easier to localize, whereas precise spatial grounding remains more challenging. Combining all three manipulations delivers the most consistent performance across benchmarks, confirming that active visual interaction fundamentally strengthens video reasoning.

\begin{table}[t]
\centering
\begingroup
\setlength{\tabcolsep}{4pt}
\setlength{\extrarowheight}{2pt}
\renewcommand{\arraystretch}{1.05}
\begin{adjustbox}{width=\columnwidth,center}
\begin{tabular}{
  c|c|c|
  >{\columncolor{blue!6}}c|
  *{5}{c}|
  >{\columncolor{blue!6}}c
}
\toprule
\rotheads{\textit{Find-Segment}} &
\rotheads{\textit{Find-Frame}} &
\rotheads{\textit{Spatial-Zoom}} &
\rotheads{\textbf{VCoM-Bench}} &
\rotheads{VidMMMU} &
\rotheads{MMVU-Val} &
\rotheads{ScienceVid} &
\rotheads{VideoMath} &
\rotheads{VideoMME} &
\rotheads{\textbf{Average}} \\
\midrule
\xxmark & \xxmark & \xxmark & 60.0 & 48.0 & 26.9 & 24.8 & 62.0 & 56.7 & 43.7 \\
\ccmark & \xxmark & \xxmark & 65.1 & 49.3 & 27.3 & 25.5 & 62.3 & 58.0 & 44.5 \\
\ccmark & \ccmark & \xxmark & 67.6 & 48.7 & 27.5 & \textbf{27.8} & 64.4 & 58.1 & 45.3 \\
\rowcolor{blue!6} \ccmark & \ccmark & \ccmark & \textbf{68.7} & \textbf{50.2} & \textbf{27.6} & \textbf{27.8} & \textbf{65.4} & \textbf{59.4} & \textbf{46.1} \\
\bottomrule
\end{tabular}
\end{adjustbox}
\endgroup
\caption{\textbf{Ablation on visual manipulations.} Each row shows which atomic manipulations (\textit{Find-Segment}, \textit{Find-Frame}, \textit{Spatial-Zoom}) are enabled. Incremental addition of these manipulations demonstrates that temporal rewatching, frame inspection, and spatial refinement collectively enhance interactive video reasoning.}
\label{tab:ablation_com}
\end{table}

\begin{table}[t]
\centering
\setlength{\tabcolsep}{3pt}
\renewcommand{\arraystretch}{1.0}
\begin{tabularx}{\columnwidth}{A{3.4cm} B B B}
\toprule
Model & \textbf{Acc.} & \textbf{Reason} & \textbf{Acc@IoU} \\
\midrule
Video-R1~\cite{feng2025videor1}   & 57.0 & 12.8 & 8.0 \\
VideoChat-R1.5~\cite{yan2025videochat_r1.5} & 59.9 & 35.1 & 28.0 \\
\midrule
Video-CoM SFT                   & 64.0 & 49.2 & 38.1 \\
Video-CoM GRPO                   & 66.7 & 51.0 & 37.6 \\
\rowcolor{blue!6} Video-CoM RA-GRPO & \textbf{68.7} & \textbf{53.8} & \textbf{39.2} \\
\bottomrule
\end{tabularx}
\caption{\textbf{Evaluation of reasoning quality on \textit{\ourbench}.}
RA-GRPO improves both reasoning accuracy and joint Acc@IoU, demonstrating its effectiveness in grounding visual evidence and maintaining answer correctness.}
\label{tab:reasoning_and_accuracy}
\end{table}

\begin{figure}[!h]
  \centering
  \includegraphics[width=\columnwidth]{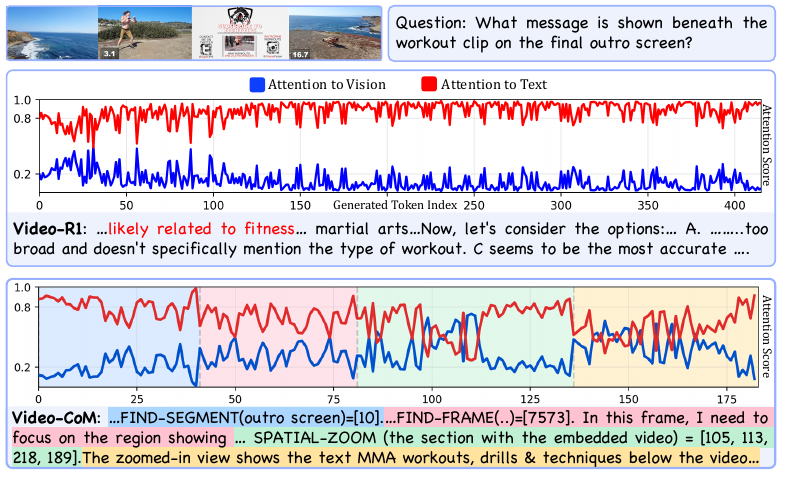}
  \vspace{-2em}
    \caption{\textbf{Comparison of visual attention.} Video-R1~\cite{feng2025videor1} (top) focuses primarily on text tokens, indicating reliance on world knowledge rather than visual evidence. \textit{Video-CoM} (bottom) maintains dynamic attention to vision tokens across reasoning rounds, demonstrating active visual reasoning.}
  \label{fig:visual_attention}
  \vspace{-1em}
\end{figure}

\noindent\textbf{Evaluation of Reasoning Quality.} Table~\ref{tab:reasoning_and_accuracy} evaluates both final answer accuracy and reasoning quality on \textit{\ourbench}. Reasoning quality denotes the fraction of trajectories with manipulation accuracy $> 0.3$, Acc@IoU jointly accounts for both reasoning precision and answer correctness. Compared to SFT and GRPO, RA-GRPO achieves consistent gains across all metrics, improving reasoning precision and accuracy. These results confirm that step-level supervision enhances the model’s ability to maintain visual grounding while preserving answer correctness.

\noindent\textbf{Analysis of Visual Attention.} To examine how models ground their reasoning, we analyze whether attention stays anchored in visual evidence or drifts toward world knowledge. As shown in Figure~\ref{fig:visual_attention}, Video-R1~\cite{feng2025videor1} focuses primarily on text tokens, indicating reliance on general associations rather than visual attention, consistent with models that \textit{think about videos}. In contrast, \textit{Video-CoM} maintains dynamic attention to vision tokens across multiple reasoning rounds (shown in distinct colors), demonstrating that the model re-engages perception at manipulation points, continually gathering and refining visual evidence. Attention maps are computed as mean last-layer attention from generated tokens to text and vision, averaged across all heads.

\vspace{-1em}
\section{Conclusion}
This work advances the goal of \textit{interactive video reasoning}, where models actively gather and refine visual evidence rather than relying on fixed video representations. We introduce \textit{\ourmodel}, which achieves this through a \textit{chain of manipulations}, supported by \textit{\ourdata}, a manipulation-driven dataset, and a \textit{reasoning-aware GRPO} objective aligning learning with step-level rewards. Despite using far less training data than prior large-scale models, \textit{\ourmodel} achieves competitive results across nine benchmarks, demonstrating the effectiveness of interactive reasoning for evidence-driven video understanding.

\section{Acknowledgement}
The computations were enabled by resources provided by the LUMI supercomputer hosted by CSC (Finland) and the LUMI consortium, and by the Berzelius resource provided by the Knut and Alice Wallenberg Foundation at the National Supercomputer Centre.

\appendix
\clearpage
\setcounter{page}{1}
\maketitlesupplementary
\part{} 
\parttoc

\section{Additional Dataset Details}
\label{appendix_sec:data}

\subsection{Video Selection from Existing Sources}
We curate the video sources for the SFT subset of \ourdata from large-scale existing datasets aggregated in Video-LLaVA-178K~\cite{zhang2024video}, which combines diverse video sources spanning human activities, instructional content, and egocentric scenes~\cite{caba2015activitynet, zhou2017youcookii, sigurdsson2016hollywood_charades, shang2019annotating, grauman2022ego4d, patraucean2023perception, yi2019clevrer}. This provides a broad base of visually and temporally varied content suitable for manipulation-driven reasoning.

Each video is accompanied by a descriptive caption used to estimate its \textit{information density}, a measure of how richly a video encodes diverse visual and temporal cues that support localized evidence gathering. When available, we utilize ground-truth or human-curated captions from the original datasets, while for other sources, captions are automatically generated using Qwen2.5-VL-72B~\cite{bai2025qwen2.5VL}.

We compute an information-density score for each video using Qwen3-235B-A22B~\cite{yang2025qwen3}, following the prompt described in Table~\ref{tab:prompt-info-density}. The scoring process evaluates the richness of visual entities, temporal diversity, and scene complexity within each caption, assigning a score in the range $[0, 10]$. Higher scores correspond to videos containing multiple temporally localized events, varied spatial layouts, and fine-grained object interactions, characteristics critical for manipulation-based reasoning.

\begin{figure}[h]
  \centering
  \includegraphics[width=0.8\columnwidth]{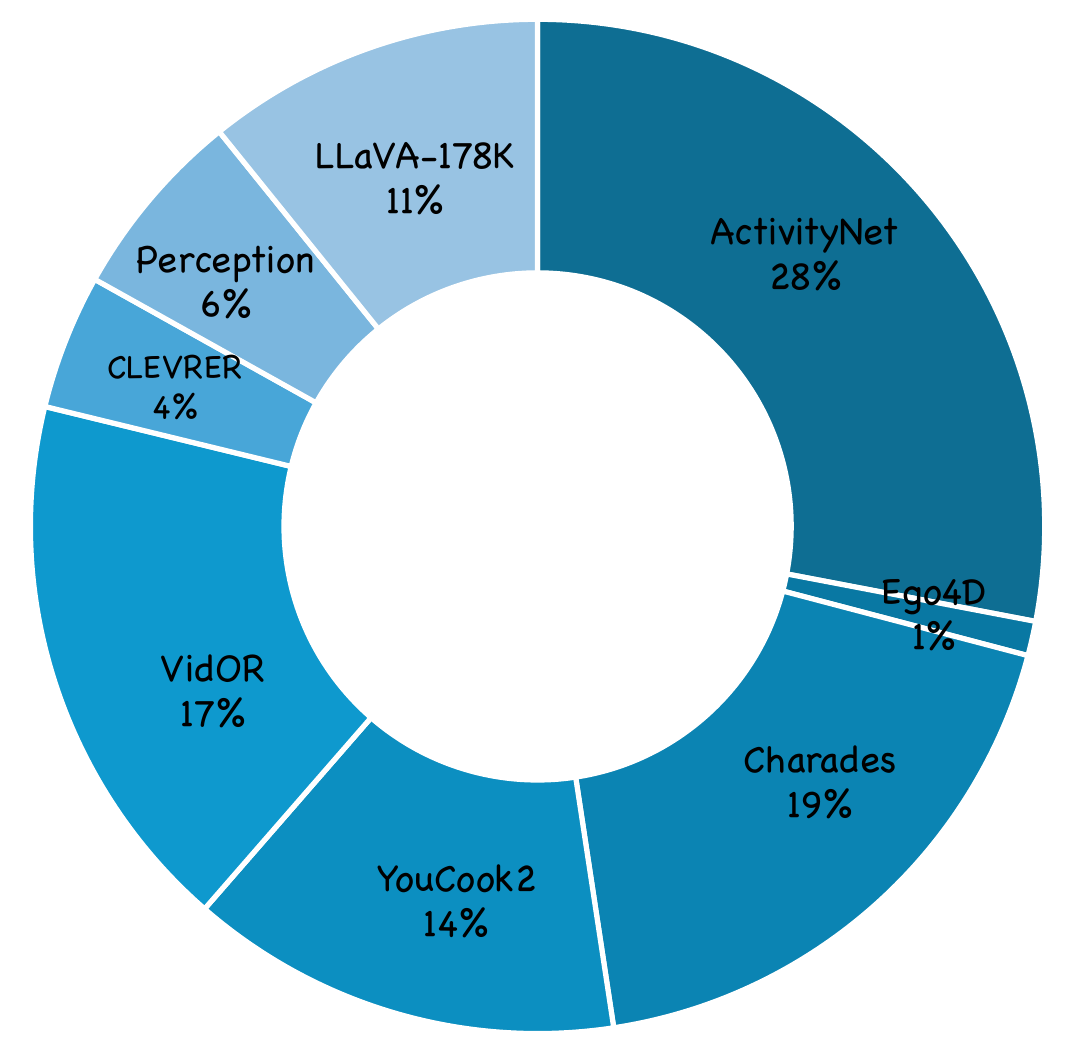}
  \caption{\textbf{Proportion of videos selected from existing sources for constructing \textit{\ourdata}.} 
  The figure shows the final composition of the $9$K videos drawn from large-scale existing datasets used in the SFT stage of \textit{\ourdata}, including ActivityNet~\cite{caba2015activitynet}, Charades~\cite{sigurdsson2016hollywood_charades}, VidOR~\cite{shang2019annotating}, YouCookII~\cite{zhou2017youcookii}, YouTube videos collected by LLaVA-178K~\cite{zhang2024video}, PerceptionTest~\cite{patraucean2023perception}, CLEVRER~\cite{yi2019clevrer}, and Ego4D~\cite{grauman2022ego4d}.}
  \label{fig:data_pie}
  \vspace{-1em}
\end{figure}

\noindent \textbf{Distribution of Videos Curated from Existing Sources.}~Based on these scores, we select approximately $9$K videos with the highest information density across all sources. This filtering process ensures that the resulting set captures both temporal diversity and spatial detail while maintaining balanced coverage across content domains. The final distribution of selected videos is illustrated in Figure~\ref{fig:data_pie}. The largest contributing sources are ActivityNet~\cite{caba2015activitynet} ($28\%$), Charades~\cite{sigurdsson2016hollywood_charades} ($19\%$), VidOR~\cite{shang2019annotating} ($17\%$), and YouCookII~\cite{zhou2017youcookii} ($14\%$), followed by YouTube videos collected by LLaVA-178K~\cite{zhang2024video}, PerceptionTest~\cite{patraucean2023perception}, CLEVRER~\cite{yi2019clevrer}, and Ego4D~\cite{grauman2022ego4d}. This diverse mixture provides a comprehensive foundation for generating high-quality, manipulation-centric question-answer pairs in the \textit{\ourdata} SFT subset.

\tcbset{
  promptbox/.style={
    enhanced,
    title filled,
    colback=gray!5!white,
    colframe=black!75!white,
    colbacktitle=blue!6,
    coltitle=black,
    fonttitle=\fontsize{11pt}{11.5pt}\selectfont\bfseries,
    toptitle=4pt,              
    bottomtitle=4pt,           
    lefttitle=4pt,             
    righttitle=4pt,
    title=Prompt for Scoring Video Information Density,
    boxrule=0.4pt,
    arc=8pt,
    outer arc=8pt,
    top=4pt, bottom=4pt, left=4pt, right=4pt,
    width=\linewidth
  }
}

\begin{table*}[t]
\centering
\begin{tcolorbox}[promptbox, colbacktitle=blue!6, coltitle=black, width=\textwidth]
You are an intelligent assistant for estimating the \textbf{information density} of a video based on its caption. 

\vspace{0.9em}

\textbf{TASK:} You will be given: a detailed text description (\texttt{caption}) summarizing the visual and temporal content of a video. Your task is to assess how much visual and temporal information this video likely contains, in terms of scene diversity, number of activities, object interactions, presence of text or small visual details, and changes over time. Higher scores correspond to videos that include multiple actions, scene transitions, or visually rich environments with fine-grained elements, while lower scores indicate static or uniform scenes with minimal variation or visual complexity.

\vspace{0.9em}

\textbf{Scoring Rubric:}
\begin{itemize}[itemsep=6pt, topsep=2pt, leftmargin=1em]
\item \textbf{9--10 (Extremely High Density)}: The caption describes complex, multi-context scenes with multiple entities and sequential actions. Such videos likely contain numerous objects, fine-grained movements, and visual changes that demand attention. \textit{Examples:} “A person assembles a drone indoors, then flies it outside while checking telemetry on a laptop”, “A reporter interviews several people while the camera pans across stalls showing different products”.

\item \textbf{7--8 (High Density)}: The caption involves several actions or interactions within a single dynamic setting. It includes diverse objects, moderate motion, and visible variations in activity or focus. \textit{Examples:} “A mechanic changes a car tire, checks pressure, and cleans the tools”, “A teacher explains a concept using diagrams while students take notes”.

\item \textbf{5--6 (Moderate Density)}: The caption includes a short sequence of simple or familiar actions. Some motion and object variation are present but limited, resulting in modest visual richness. \textit{Examples:} “A person pours milk into coffee and stirs”, “A dog chases a ball and returns it”.

\item \textbf{3--4 (Low Density)}: The caption depicts a mostly static scene with limited activity or change, few objects or movements. \textit{Examples:} “A woman applying makeup facing the camera”, “Close-up shot of someone feeding a bird”.

\item \textbf{1--2 (Very Low Density)}: The caption describes a nearly static or visually uniform scene with no significant motion, object variation, or scene change. \textit{Examples:} “A sunset over the mountains”, “A close-up of a flower”.
\end{itemize}

\vspace{0.9em}
\textbf{Evaluation Principles:} Favor higher scores for captions describing multiple activities, diverse objects, visible text or small details, noticeable scene transitions, and temporal changes. Favor lower scores for static or repetitive scenes with minimal motion, uniform backgrounds, or limited visual elements. Base the assessment strictly on the caption without inferring additional information.

\vspace{0.9em}
\textbf{Output Format}\\
\noindent\texttt{SCORE\_CARD: \{"critique": "<2-3 sentence justification summarizing scene diversity, activity variety, and visual-temporal richness>", "score": <1-10>\}}

\end{tcolorbox}
\caption{Prompt for scoring video information density. Higher scores correspond to videos with greater scene diversity, multiple actions, object interactions, and fine-grained visual details.}
\label{tab:prompt-info-density}
\end{table*}
\tcbset{
  promptbox/.style={
    enhanced,
    title filled,
    colback=gray!5!white,
    colframe=black!75!white,
    colbacktitle=blue!6,
    coltitle=black,
    fonttitle=\fontsize{11pt}{11.5pt}\selectfont\bfseries,
    toptitle=4pt,              
    bottomtitle=4pt,           
    lefttitle=4pt,             
    righttitle=4pt,
    title=Prompt for Question-Answer Generation in \textit{\ourdata},
    boxrule=0.4pt,
    arc=8pt,
    outer arc=8pt,
    top=4pt, bottom=4pt, left=4pt, right=4pt,
    width=\linewidth
  }
}
\begin{table*}[!t]
\vspace{-0.5em}
\centering
\begin{tcolorbox}[promptbox, colbacktitle=blue!6, coltitle=black, width=\textwidth]
You are given a video. Your task is to create ten challenging question-answer pairs that require close inspection of the visual content using the defined tools. These questions should not be easily answered by simply watching the video from beginning to end one time. The solver should be required to rewatch specific portions, pause at precise frames, or zoom into small regions to find the answer.

\vspace{0.1em}

\medskip
\textbf{TARGET:} Challenging or difficult questions can come from either of the following types:
\begin{itemize}[itemsep=1pt, topsep=2pt, leftmargin=1em]
\item Locating the right segment in a long video and rewatching it to find the answer.
\item Freezing the exact frame to observe objects, actions, or details (object presence, colour, counting, order, subtle actions).
\item Zooming into small regions to check for clear text, logos, numbers, or tiny objects (text or any fine details).
\item Combinations of the above three types.
\end{itemize}

\vspace{0.4em}

\textbf{TOOLS:}
The video is divided into equal-length segments. Segment and frame identifiers are overlaid in the top-left corner. Audio is unavailable, so questions must not rely on names or spoken dialogue.
\begin{enumerate}[itemsep=3pt, topsep=2pt, leftmargin=1.2em]
\item \textbf{Find-Segment}: The video is divided into segments. This tool lets the solver rewatch a specific segment of the video in more detail. It is essential when the video is long (has multiple segments) or when events unfold over time. Unlike a single-frame detail, segment-based questions depend on understanding the action that takes place in a segment or progression of actions or the order in which events happen. For example, asking “Which object is picked up first before the person leaves the room?” or “How many people does the man shake hands with?” forces the solver to locate and carefully rewatch that segment. Such questions cannot be solved by only looking at one frame. A question targets rewatching one or a maximum of two segments for such reasoning, never more, so that exploration remains focused but still requires deliberate navigation. Indirectly this should test the ability to locate the most appropriate segment to answer the question.

\item \textbf{Find-Frame}: This tool freezes the video at a specific frame. It is required when the answer depends on a clear visual observation that can be made certain by looking at a specific frame: counting objects, identifying objects or outfits, recognizing an attribute such as colour, or checking whether an object is present. \textit{Find-Frame} can be used on its own if it is easy to locate the frame, or after \textit{Find-Segment} when the solver must first navigate to the correct part of the video. The key purpose of this tool is to extract details more confidently that cannot be identified correctly by simply skimming or recalling the overall video.

\item \textbf{Spatial-Zoom}: This tool zooms into a specific region of a frame to reveal small or distant details that are otherwise unreadable or unclear. It is especially useful for text on signs, logos, numbers on jerseys, or small objects hidden within cluttered scenes. Without zooming in, such details would remain either invisible or easily misinterpreted. \textit{Spatial-Zoom} is always used after \textit{Find-Frame} (to first stop at the right moment). For example, to read text on a bottle label or to check a tiny object in someone's hand, the solver must freeze the correct frame and then zoom in on the region of interest. This ensures that the answer comes from precise inspection, not broad guessing.
\end{enumerate}

\vspace{0.5em}

\textbf{TASK}: Write 10 questions that cannot be answered correctly without using the above tools. For each question: Write the question, correct short answer, 5 multiple-choice options (A-E) with exactly one correct option and provide a brief reasoning plan listing the sequence of tools.

\vspace{0.5em}

\textbf{Example Reasoning Paths}
\begin{itemize}[itemsep=1pt, topsep=2pt, leftmargin=1em]
\item Only \textit{Find-Segment}: when the answer depends on actions or the order of events within a segment.
\item Only \textit{Find-Frame}: when the answer depends on a clear, stable visual detail or counting in a single frame.
\item \textit{Find-Segment}→ \textit{Find-Frame}: relevant detail is brief and requires zooming to a segment before freezing.
\item \textit{Find-Frame}→ \textit{Spatial-Zoom}: when the needed detail is small or distant and must be read or inspected after freezing.
\item \textit{Find-Segment}→ \textit{Find-Frame}→ \textit{Spatial-Zoom}: when the answer requires segment selection, precise frame selection, and zooming for small details.
\end{itemize}

\vspace{0.5em}

\textbf{Examples}: \texttt{Insert the in-context examples here.}

\end{tcolorbox}
\vspace{-0.5em}
\caption{\textbf{Prompt for QA generation in \textit{\ourdata}}.
The prompt defines a structured framework for creating interactive video reasoning questions that require explicit use of manipulations. It instructs the intended purpose and use cases of each operation, guiding the model to generate questions that demand targeted visual inspection and reasoning trajectories.}
\label{tab:data_qa_generation_prompt}
\end{table*}

\subsection{Question-Answer Generation}

Building upon the $9$K videos sampled from existing sources and the $70$ manually curated high-complexity videos, we generate the initial pool of question-answer (QA) pairs for interactive video reasoning. Each video serves as a source for $3$-$5$ QA pairs, resulting in approximately $25$K candidate examples prior to filtering. The objective of this stage is to construct questions that explicitly elicit manipulation-based reasoning, requiring the model to actively explore the video to gather visual evidence rather than rely on global context.

However, for the manually curated videos, we generate and retain up to $20$ QA pairs per video. These videos are intentionally selected to provide rich temporal and spatial complexity, offering multiple opportunities to form manipulation-specific questions.  

The prompt used with Gemini-2.5-Pro~\cite{comanici2025gemini} to generate QA pairs is provided in Table~\ref{tab:data_qa_generation_prompt}. The prompt establishes a structured framework for producing interactive video reasoning questions by explicitly defining the available manipulation tools, \textit{find-segment}, \textit{find-frame} and \textit{spatial-zoom}, and describing their appropriate use cases. It instructs the model not only to create questions requiring these tools but also to articulate reasoning paths that reflect deliberate visual exploration. Each question must therefore demand targeted inspection: identifying the correct segment, isolating precise frames, or zooming into fine spatial details. The prompt further specifies representative reasoning paths (e.g., \textit{find-segment}$\rightarrow$\textit{find-frame}$\rightarrow$\textit{spatial-zoom}) to encourage coverage across temporal, spatial, and fine-grained reasoning levels, ensuring that the generated QA pairs comprehensively span the manipulation space.

\subsection{Reasoning Generation and Verification}
After generating the initial QA pairs, we perform a two-stage verification process to ensure both correctness and reasoning quality. Each question is re-evaluated by Gemini-2.5-Pro~\cite{comanici2025gemini} in two complementary formats: multiple-choice (MCQ) and open-ended. For each format, the model produces a reasoning trace followed by a final answer.

This process serves two primary purposes. First, the reasoning trace provides an explicit step-by-step solution describing the sequence of manipulations, thereby enriching each sample with structured reasoning supervision. Second, by comparing the model’s final prediction against the ground-truth answer in both formats, we filter out ambiguous, inconsistent, or incorrect samples. Only examples where both predictions match the correct answer are retained in the final dataset.

The prompt used for this step, shown in Table~\ref{tab:data_qa_verification_prompt}, provides a detailed framework for solution generation through active video reasoning. It defines that each question must be solved using the atomic manipulations and specifies when and why each operation should be invoked. The prompt further outlines the different reasoning trajectories, ensuring that solutions follow interpretable and compositional chains of manipulations rather than text-only reasoning.

\subsection{Structured Reasoning}  
Importantly, the response format enforces a structured reasoning trace that interleaves three complementary reasoning components.
\begin{itemize}
    \item \textbf{Exploratory-Reasoning} performs textual reasoning conditioned on the currently available visual information to decide \textit{what} manipulation to perform next and \textit{where} to focus within the video. It bridges perception and action by hypothesizing the most relevant temporal or spatial region for evidence gathering.
    \item \textbf{Visual-Manipulation} executes the manipulation to perform the corresponding visual action to gather localized evidence.
    \item \textbf{Observation} describes and uses the resulting new visual observation for continued reasoning.
\end{itemize}

\begin{figure*}[!t]
\centering
\includegraphics[width=\textwidth]{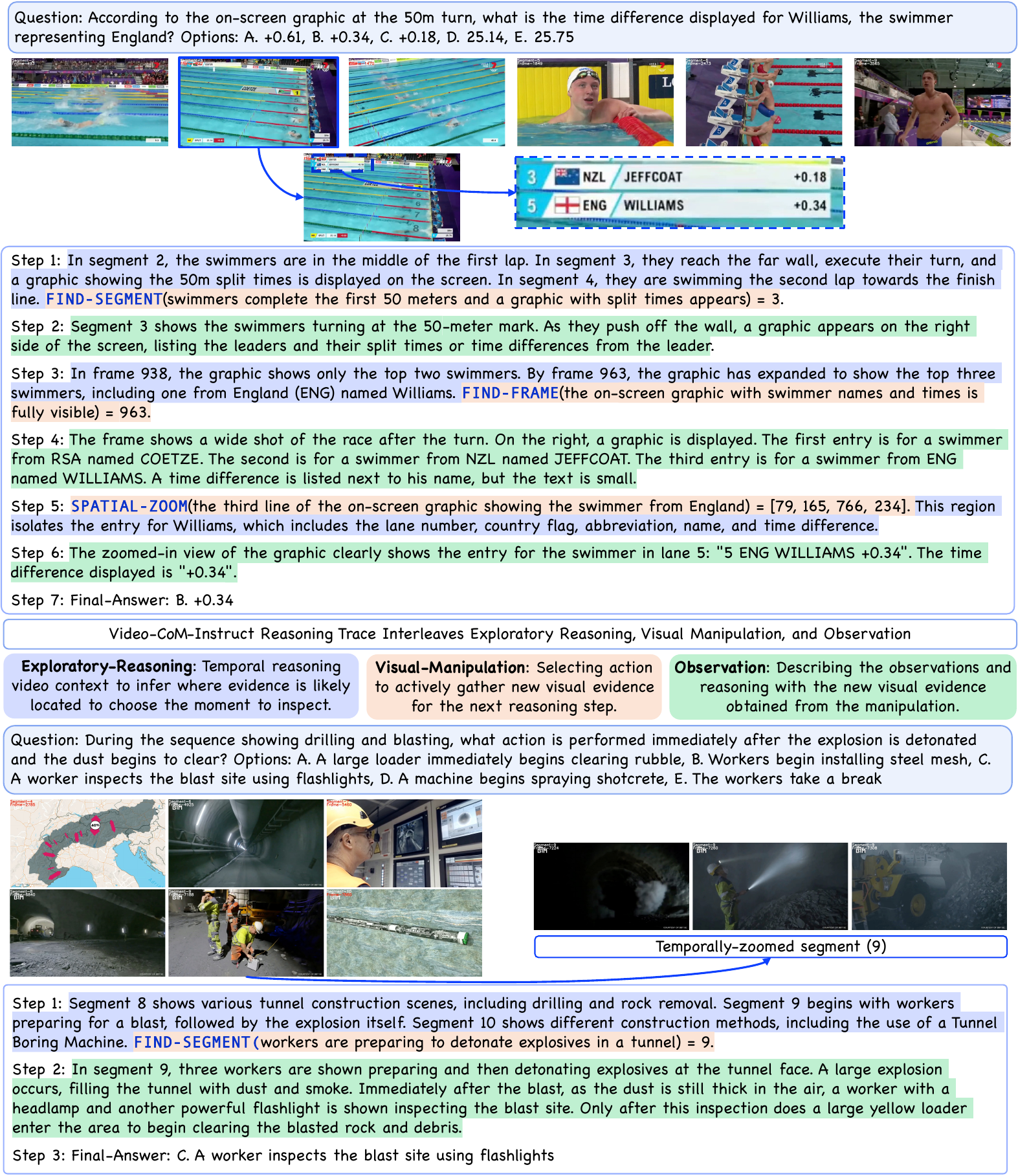}
\vspace{-1em}
\caption{\textbf{Reasoning traces in the \textit{\ourdata} dataset interleave exploratory-reasoning, visual-manipulation, and observation.}
\textit{Top}: Temporal zooming is applied to capture the brief moment when the on-screen timing graphic becomes fully visible, followed by a spatial zoom to read the fine-grained split times.
\textit{Bottom}: A fleeting inspection during a tunnel explosion sequence, where the reasoning trace focuses on the short temporal window and observes the worker inspecting the blast site with a flashlight. These examples illustrate reasoning structured as exploratory-reasoning → visual-manipulation → observation.}
\label{fig:data_example_1}
\end{figure*}

As a result, each reasoning trace in \textit{\ourdata} follows this structured progression, as illustrated in Figure~\ref{fig:data_example_1}. Each manipulation is therefore immediately followed by an observation that reports what has been discovered, verified, or localized. The format specifies clear localization criteria, requiring the model to articulate distinct visual cues (objects, motions, colors, text, or spatial arrangements) that justify each step and distinguish the chosen segment, frame, or region from neighboring contexts. This enforced alternation of “\textbf{Exploratory-Reasoning → Visual-Manipulation → Observation},” ensures that the reasoning trace unfolds as a sequence of \textit{verifiable perception-action pairs}, where every claim in the textual reasoning is grounded in direct visual evidence rather than abstract inference.

Finally, the prompt also requires the model to produce dense spatio-temporal annotations in the form of \textit{Valid-Segments} and \textit{Valid-Frames}, marking all temporally or spatially valid regions where evidence supporting the answer appears. These annotations are later leveraged during RA-GRPO training to compute step-level reasoning rewards, directly aligning policy optimization with the annotated reasoning trajectories.

\tcbset{
  promptbox/.style={
    enhanced,
    title filled,
    colback=gray!5!white,
    colframe=black!75!white,
    colbacktitle=blue!6,
    coltitle=black,
    fonttitle=\fontsize{11pt}{11.5pt}\selectfont\bfseries,
    toptitle=4pt,              
    bottomtitle=4pt,           
    lefttitle=4pt,             
    righttitle=4pt,
    title=Prompt for Step-by-step Solution Generation and Verification,
    boxrule=0.4pt,
    arc=8pt,
    outer arc=8pt,
    top=4pt, bottom=4pt, left=4pt, right=4pt,
    width=\linewidth
  }
}

\begin{table*}[!t]
\centering
\begin{tcolorbox}[promptbox, colbacktitle=blue!6, coltitle=black, width=\textwidth]
You are given a video and a multiple-choice question. Your task is to produce a clear, step-by-step solution for the question using the defined tools, and then give the final answer that exactly matches one of the options. Your solutions must involve active inspection of the visual content by locating segments, freezing frames, and zooming into regions.

\medskip
\textbf{TARGET:} The questions are generated to target solutions that require active visual inspection in the following ways:
\begin{itemize}[itemsep=1pt, topsep=2pt, leftmargin=1em]
\item Locating the right segment in a long video and rewatching it to find the answer.
\item Freezing the exact frame to observe objects, actions, or details (object presence, colour, counting, order, subtle actions).
\item Zooming into small regions to check for clear text, logos, numbers, or tiny objects (text or any fine details).
\item Combinations of the above three types.
\end{itemize}

\medskip
\textbf{TOOLS}: The video is divided into equal-length segments. Segment and frame IDs are overlaid in the top-left corner. The following tools allow you to actively explore and inspect the video to find the answer, instead of passively watching.

\begin{enumerate}[itemsep=1pt, topsep=2pt, leftmargin=1.2em]
\item \textbf{Find-Segment}: The video is divided into segments. This tool lets the model rewatch a specific segment of the video in more detail. It is essential when the video is long (has multiple segments) or when events unfold over time. Unlike a single-frame detail, segment-based answers depend on understanding the progression of actions and the order in which events happen. For example, solving “Which object is picked up first before the person leaves the room?” requires locating and carefully rewatching that segment. Such answers cannot be derived by only looking at one frame. Use one or a maximum of two segments for such reasoning, never more, so that exploration remains focused but still requires deliberate navigation. Indirectly this proves the ability to locate the most appropriate segment to solve the question.

\item \textbf{Find-Frame}: This tool freezes the video at a specific frame. It is required when the answer depends on a clear visual observation that can be made certain by looking at a specific frame: counting people sitting on a bench, identifying which object is being held or an outfit, recognizing an attribute such as colour, or checking whether an object is present. \textit{Find-Frame} can be used on its own if it is easy to locate the frame, or after \textit{Find-Segment} when the model must first navigate to the correct part of the video. The key purpose of this tool is to extract details that cannot be identified correctly by simply skimming or recalling the overall video, ensuring the answer comes from precise inspection.

\item \textbf{Spatial-Zoom}: This tool zooms into a specific region of a frame to reveal small or distant details that are otherwise unreadable or unclear. It is especially useful for text on signs, logos, numbers on jerseys, or small objects hidden within cluttered scenes. Without zooming in, such details would remain either invisible or easily misinterpreted. \textit{Spatial-Zoom} is always used after \textit{Find-Frame} (to first stop at the right moment). For example, to find the solution to “What is written on a bottle label?” the model must freeze the correct frame and then zoom in on the region of interest. This ensures that the answer comes from precise inspection, not broad guessing.
\end{enumerate}

\medskip
\textbf{Possible Reasoning Paths}: 
\begin{enumerate}[itemsep=1pt, topsep=2pt, leftmargin=1em]
\item Only \textit{Find-Segment}: used when the answer depends on actions within a segment, such as what someone does or the order in which two things happen. Usually only one or two segments need to be rewatched.
\item Only \textit{Find-Frame}: used when a clear detail can be seen long enough to freeze the video at the right moment. This can include checking what object someone is holding, counting how many people or items are present, identifying the color or pattern of clothing, reading large or clearly visible text, or locating the position of an object in the scene. The key idea is that the answer comes from careful attention to one frame rather than following the whole sequence.
\item \textit{Find-Segment} → \textit{Find-Frame}: used when the detail is short in duration, so the model first needs to narrow down to the correct segment, then carefully look at its frames to freeze the exact moment.
\item \textit{Find-Frame} → \textit{Spatial-Zoom}: used when the detail is visible in a frame but too small or distant to be read or recognized without zooming. This includes reading fine print on a sign, checking a small logo on clothing, identifying a small object, or inspecting a distant detail in the background. The model must first stop at the right frame and then zoom into the specific region.
\item \textit{Find-Segment} → \textit{Find-Frame} → \textit{Spatial-Zoom}: used when the model has a rough idea that the answer lies within a segment but must then pause on the exact frame and zoom in to uncover the detail. This is useful when an event happens within a known segment, but the detail itself is small or not visible without zooming.
\end{enumerate}
(Continued...)
\end{tcolorbox}
\end{table*}

\tcbset{
  promptbox/.style={
    enhanced,
    title filled,
    colback=gray!5!white,
    colframe=black!75!white,
    colbacktitle=blue!6,
    coltitle=black,
    fonttitle=\fontsize{11pt}{11.5pt}\selectfont\bfseries,
    toptitle=4pt,              
    bottomtitle=4pt,           
    lefttitle=4pt,             
    righttitle=4pt,
    title=Prompt for Step-by-step Solution Generation and Verification (Cont.),
    boxrule=0.4pt,
    arc=8pt,
    outer arc=8pt,
    top=4pt, bottom=4pt, left=4pt, right=4pt,
    width=\linewidth
  }
}
\begin{table*}[!t]
\vspace{-0.7em}
\centering
\begin{tcolorbox}[promptbox, colbacktitle=blue!6, coltitle=black, width=\textwidth]
\medskip
\textbf{TASK:} For each question:
\begin{itemize}[itemsep=1pt, topsep=2pt, leftmargin=1em]
\item Read the question and options, follow the most appropriate reasoning path (or the suggested one if suitable), record brief observations after each tool use, and conclude with a final answer that exactly matches one of the option texts.
\item Provide \textit{Valid-Segments} and \textit{Valid-Frames} when segment or frame tools are used.
\end{itemize}

\medskip
\textbf{Important Reasoning Rules}
\begin{itemize}[itemsep=1pt, topsep=2pt, leftmargin=1em]
\item Do not use timestamps; instead, navigate by events, visual cues, segments, and frames.
\item For \textit{Spatial-Zoom}, the phrase must be clear and descriptive enough to uniquely ground the target object or region based only on the current frame.
\item \textit{Valid-Segments} and \textit{Valid-Frames} must densely list every possible location where the decisive evidence is clearly visible. If any portion of a segment contains the evidence, include the entire segment ID. For frames, include the full continuous range where the evidence remains visible and stable (expand a few frames before and after the chosen one), but do not include frames where the evidence disappears or is obstructed. Only include \textit{Valid-Segments} if a segment was used, and \textit{Valid-Frames} if a frame was used.
\end{itemize}

\medskip
\textbf{OUTPUT FORMAT}
\medskip

\textbf{Example for \textit{Find-Segment} → \textit{Find-Frame} → \textit{Spatial-Zoom}}  
\begin{itemize}[itemsep=1pt, topsep=2pt, leftmargin=1.2em]
    \item {Step 1:} \textit{{Find-Segment}(\textless brief cue\textgreater)} = $[s]$ or $[s_1, s_2]$. Provide 2–3 concrete visual cues (actions, objects, outfits, background, text, or positions; choose a mix of the most contrastive visual details) that clearly distinguish this segment from nearby ones, proving correct temporal localization. The segment number is overlaid in the top-left corner and must be precisely accurate. Localize the correct segment required to extract the answer.
    
    \item {Step 2:} List clear observations from rewatching the chosen segment(s): what is happening, who or what, where, and key cues. (Mandatory to include if \textit{Find-Segment} is used.)
    
    \item {Step 3:} \textit{{Find-Frame}(\textless what you freeze for or why\textgreater)} = $[f]$ (or a small set like $[f_1, .., f_2]$ if helpful). Provide 2–3 concrete visual cues (actions, objects, outfits, background, text, or positions; choose a mix of the most contrastive visual details) that clearly distinguish this frame from nearby ones, proving correct frame localization. The frame number is overlaid in the top-left corner and must be precisely accurate.
    
    \item {Step 4:} List precise visual information that is visible in the selected frame. (MUST include if \textit{Find-Frame} is used.)
    
    \item {Step 5:} \textit{{Spatial-Zoom}(\textless region or what detail you want\textgreater)} = $[x_1, y_1, x_2, y_2]$. Provide 2–3 concrete visual cues that clearly describe the location of this region in the frame, proving correct spatial localization.
    
    \item {Step 6:} Write clear observations from the zoomed region: exact text, logo, number, or object you can now read or identify. (Mandatory to include if \textit{Spatial-Zoom} is used.)
    
    \item {Step 7:} \textit{Final-Answer: \textless exact option text\textgreater}
\end{itemize}
\medskip

\noindent\textit{{Valid-Segments:}} $[s_1, s_2]$ (All possible segments that can lead to the correct answer) \\
\textit{{Valid-Frames:}} $[f_1start - f_1end; f_2start - f_2end]$ (All possible ranges of frames that can lead to the correct answer)

\medskip
\textbf{EXAMPLES}: \texttt{Insert the in-context examples here}
\end{tcolorbox}
\vspace{-0.5em}
\caption{\textbf{Prompt for reasoning generation and verification in \textit{\ourbench}.} The prompt provides a structured template for generating reasoning traces through active video interaction. It enforces the use of defined manipulations and specifies when and why each should be applied, producing interpretable reasoning trajectories that integrate exploratory reasoning, visual-manipulation, and observation.}
\label{tab:data_qa_verification_prompt}
\end{table*}
\subsection{Limitations and Challenges}

Despite the careful design of Video-CoM-Instruct, constructing high-quality data for manipulation-based video reasoning presents several challenges.

\noindent \textbf{Spatial Localization in Videos.} The first and most significant limitation lies in the accuracy of spatial localizations within videos. Generating realistic questions that reflect real-world use cases, such as identifying the price of an item in Figure~\ref{fig:method_fig}, often requires reasoning over textual regions rather than clearly bounded objects. Two factors contribute to this difficulty.
\textit{(i)} Spatial localization directly within videos remains a persistent challenge even for state-of-the-art multimodal LLMs such as Gemini-2.5-Pro~\cite{comanici2025gemini} and InternVL3-78B~\cite{zhu2025internvl3}. Accurately grounding a region in the spatio-temporal domain of a video requires large-scale, high-quality supervision, which is scarce at present.
\textit{(ii)} Localizing regions containing texts or numbers is inherently more difficult than object localization. Even with static image inputs, these targets fall outside the common segmentation or detection vocabularies used by current models, leading to frequent failures. Consequently, scaling high-quality spatial localization data for videos is non-trivial. To ensure reliable supervision for reinforcement learning, we therefore manually annotate spatial regions in the GRPO subset of the dataset, but this limits the scalability of data.

\noindent \textbf{Video Sources for Manipulation-Specific Data.} A second challenge arises from the dependence of manipulation-specific data on video context. Videos with limited scene diversity, such as a single-view recording of a person cooking, offer little variation across time. In such cases, questions can often be answered without iterative visual interaction, since relevant details persist throughout the clip. This makes it difficult to construct samples that genuinely require temporal revisitation, frame inspection, or spatial zooming. To address this, we manually curate videos that exhibit richer contextual variation, ensuring the dataset includes scenarios where interactive manipulation is both necessary and beneficial for reasoning. However, this manual curation also constrains large-scale data expansion, as sourcing diverse, information-dense videos with manipulation-relevant context remains resource-intensive.

\section{Additional Implementation Details}
\label{appendix_sec:implementation}

\subsection{Video Processing.}
All videos are uniformly sampled at 2 FPS. A maximum of 32 frames are used during both supervised fine-tuning (SFT) and GRPO training, and evaluation likewise operates with up to 32 frames for all reasoning models. Each frame is resized to a maximum resolution of 60$\times$420\,px during both training and inference. 

To support temporal reasoning, explicit segment and frame indices are overlaid onto every frame. Videos are partitioned into uniform temporal segments whose durations are dynamically adjusted to fall within an adaptive range of approximately 10–30 seconds, ensuring consistent granularity across videos of varying length. Each frame is annotated in the format \texttt{Segment-{i}} and \texttt{Frame-{j}}, indicating its segment and frame positions within the overall timeline. The annotations are consistently positioned on each frame, with font size scaled relative to resolution and color adapted to local background to ensure readability.

\noindent
\subsection{Stage I - Supervised Fine-Tuning}
SFT is performed using DeepSpeed ZeRO-3 optimization, adopting \texttt{bf16} precision and a maximum context length of 32K tokens. We use a global batch size of 32, and optimize with a learning rate of $1\times10^{-5}$ under a cosine decay schedule with a 3\% warm-up ratio.The maximum number of reasoning turns is five, corresponding to the longest reasoning trajectory in the dataset.

\noindent\textbf{Data-Mix}: The base model, {Qwen2.5-VL-7B-Instruct}~\cite{bai2025qwen2.5VL}, is fine-tuned using three datasets: 9K samples from ActivityNet~\cite{caba2015activitynet} for temporal reasoning, 15K samples from the \textit{\ourdata}-SFT subset (concatenated twice, yielding 30K effective samples) for manipulation reasoning, and 180K samples from {Visual-CoT}~\cite{shao2024visual} for spatial understanding. To maintain comparable update counts between modalities despite dataset size differences, video datasets (ActivityNet and \textit{\ourdata}) are trained with a batch size of 4, while {Visual-CoT} is trained with a larger batch size of 16, ensuring balanced iteration frequencies between image and video streams within the same training schedule.

\noindent\textbf{ActivityNet for Temporal Localization}: For the ActivityNet temporal reasoning subset, timestamp-based annotations are converted into frame-level supervision to enable fine-grained temporal localization. Each training instance consists of a natural-language question referring to a specific event and an answer grounded in explicit frame indices. For example:
\textit{Q: When do the man and woman stop their individual performances and come together in applause?}
\textit{A: Frame 314 to 348, The man and woman stop their individual performances and come together in applause between Frame 314 and 348. During this period, the woman stops dancing and begins clapping, while the man stops playing the bongos and gives a final clap.}
This conversion provides dense frame-level reasoning supervision, enhancing the model’s ability to align textual cues with precise temporal boundaries throughout video sequences.

\begin{table}[!t]
\centering
\resizebox{\linewidth}{!}{
\begin{tabular}{lll}
\toprule
\textbf{Stage} & \textbf{Setting} & \textbf{Value / Notes} \\
\midrule

\multirow{6}{*}{{Video Setup}} 
& FPS & 2 (uniform sampling) \\
& Frame Sampling & 32 (train / eval) \\
& Resolution & 360$\times$420\,px \\
& Segment duration & 10–30\,s adaptive \\
& Segment Sampling & 8 (sampled uniformly) \\

\midrule
\multirow{7}{*}{{SFT}} 
& Model & Qwen2.5-VL-7B-Instruct \\
& Learning rate & $1\times10^{-5}$ (cosine) \\
& Global batch & 32 \\
& Epochs & 1 \\
& Context length & 32K tokens \\
& Max turns & $\leq$5 \\
& Batch ratio & 4 (video) / 16 (image) \\

\midrule
\multirow{7}{*}{{RA-GRPO}} 
& Global batch & 64 \\
& Learning rate & $1\times10^{-6}$ \\
& Group size & 8 \\
& KL coeff & 0.04 \\
& Rewards & Accuracy + Reasoning \\
& IoU thresh ($\tau_b$) & 0.3 (spatial) \\
& Max turns & $\leq$5 \\

\bottomrule
\end{tabular}
}
\caption{Summary of key hyperparameters used in training \textit{\ourmodel}. The table reports core configuration settings for video preprocessing, supervised fine-tuning (SFT), and reasoning-aware GRPO (RA-GRPO) training, including frame sampling, learning rates, batch sizes, and reward-related parameters.}
\label{tab:hyperparams}
\end{table}

\subsection{Stage II - Reasoning Aware GRPO}
In the second stage, we fine-tune the model using \textit{Reasoning-Aware GRPO} (RA-GRPO) on the 3K GRPO subset of \textit{\ourdata}. Training is performed on 8 GPUs with a global batch size of 64, using a learning rate of $1\times10^{-6}$ and the ZeRO-3 optimizer in \texttt{bf16} precision. Each input prompt produces $8$ candidate generations per sample to estimate group-based policy gradients. A KL-anchor regularization coefficient of $0.04$ is applied to stabilize optimization around the supervised policy. The total reward combines two components: an \textit{accuracy reward} measuring the correctness of the final answer and a \textit{reasoning reward} capturing step-level alignment between generated reasoning chains and annotated manipulation steps. For spatial localization in \textit{spatial-zoom} manipulation, we define a correctness threshold $\tau_b=0.3$ based on the intersection-over-union (IoU) between predicted and ground-truth regions. During GRPO, the maximum number of reasoning turns is capped at five. Reasoning is terminated early if the model reaches this limit without producing a final answer. For find-segment manipulations, we uniformly sample 8 additional frames from the selected segment to provide sufficient temporal context for subsequent reasoning steps. A detailed summary of all key hyperparameters for both the training stages and the evaluation is provided in Table~\ref{tab:hyperparams}.

\noindent \textbf{Training Tricks for Learning Manipulations.}~To ensure that the model effectively learns to perform and localize manipulations, we design a targeted sampling strategy during GRPO optimization. For \textit{find-segment} and \textit{find-frame}, where dense ground-truth annotations are available, at least 4 frames are sampled from within the valid temporal or frame ranges. When a \textit{find-segment} operation is immediately followed by a \textit{find-frame}, we explicitly exclude the target frame from the segment-level sampling so that the model learns the intended manipulation trajectory, first locating the correct segment, then identifying the key frame.

For \textit{find-frame} manipulations, we likewise ensure that at least four frames are drawn from the valid frame-level ground-truth region to guarantee the model observes the relevant evidence. When \textit{find-frame} follows find-segment, its sampled frames are restricted to those obtained from the corresponding segment; otherwise, frames are drawn from the initial 32 globally sampled frames. This controlled sampling procedure ensures that the model is consistently exposed to the relevant visual evidence required to identify correct manipulations, while maximizing the effective utilization of each training sample during GRPO optimization.

\subsection{Inference Efficiency and Latency}
To assess the computational efficiency of our approach, we measure the average inference latency of \textbf{Video-CoM} and recent baselines during evaluation on \textbf{VCoM-Bench}. All models are evaluated on a single \texttt{AMD Instinct\texttrademark\ MI210} GPU with 64\,GB memory under the same 2\,FPS, 32-frame setting described in the main text. Latency is defined as the wall-clock time per iteration, averaged over the complete benchmark evaluation. 

Although our Video-CoM uses up to five reasoning turns, its latency increases only moderately compared with single-turn Video-R1 and three-iteration VideoChat-R1.5. This is because Video-CoM reuses encoded visual features across turns rather than re-encoding the video at each step. Further, Video-R1 tends to produce longer reasoning sequences within a single iteration, resulting in higher overall latency despite single turn. The results indicate that interactive multi-turn reasoning achieves reasonable accuracy gains while maintaining practical inference efficiency.

\begin{table}[!h]
\centering
\resizebox{\columnwidth}{!}{%
\begin{tabular}{lccc}
\toprule
\textbf{Model} & \textbf{Frames} & \textbf{Latency (s)} & \textbf{Accuracy (\%)} \\
\midrule
Video-R1~\cite{feng2025videor1} & 32 & 19.04 & 57.1 \\
VideoChat-R1.5~\cite{yan2025videochat_r1.5} & 32 & 20.27 & 59.9 \\
\rowcolor{blue!6}
\textbf{Video-CoM (Ours)} & 32 & 25.90 & 68.7 \\
\bottomrule
\end{tabular}%
}
\caption{Inference efficiency comparison on VCoM-Bench. Latency is measured as the average wall-clock time per iteration on a single \texttt{AMD Instinct\texttrademark\ MI210} GPU with 64\,GB memory. All models use a maximum of 32 frames. Video-R1 uses single turn, VideoChat-R1.5 uses up to three iterations and our VVideo-CoM uses up to 5 iterative turns during evaluation.}
\label{tab:vcom_efficiency}
\end{table}
\section{Additional Benchmark Details}

\begin{figure*}[!t]
  \centering
  \includegraphics[width=\textwidth]{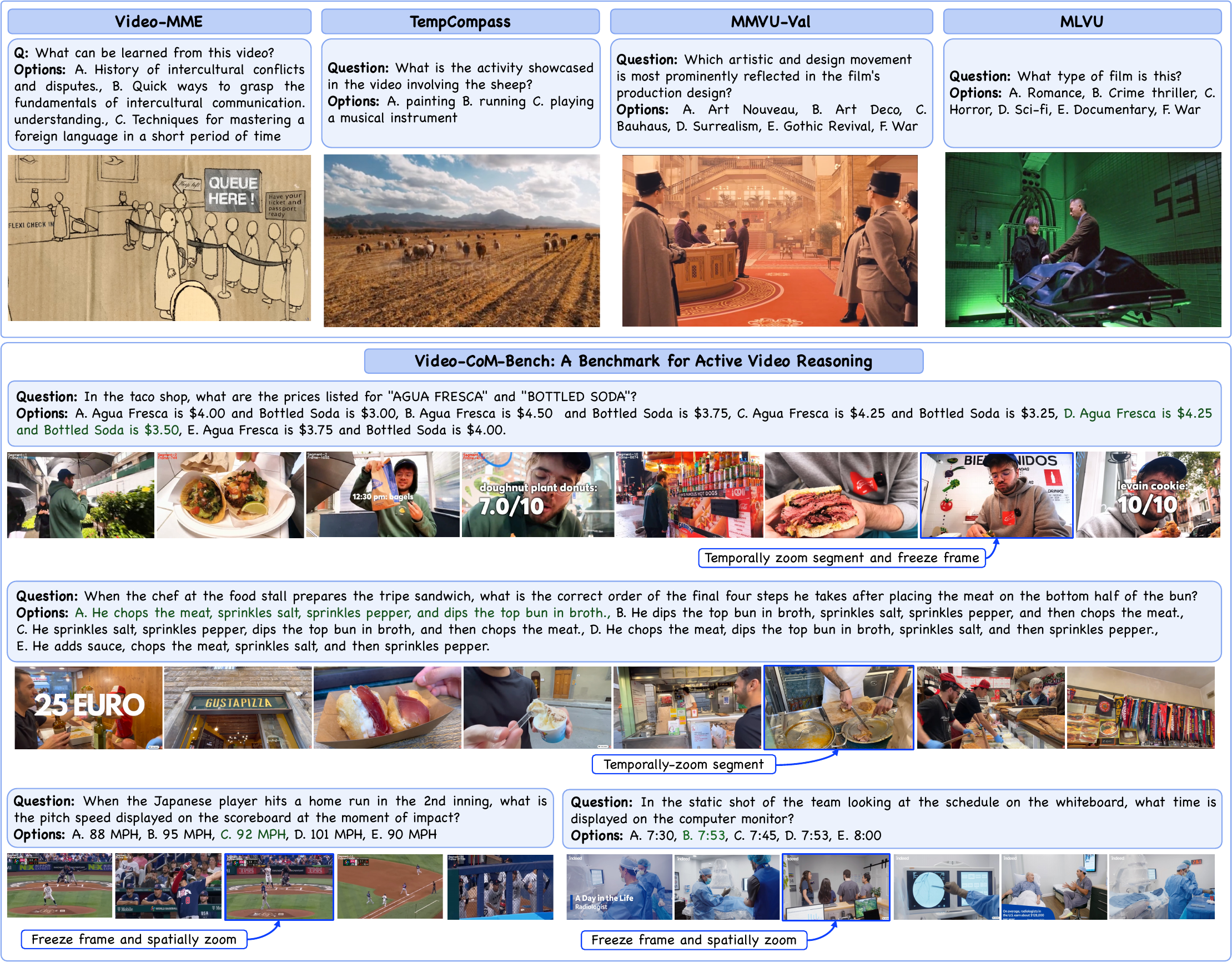}
  \caption{\textbf{Benchmarking the Ability to Think with Videos.}
Existing benchmarks~\cite{li2023videomme, tempcompass, mmvu2024, zhou2025mlvu}~(top) primarily focus on holistic video understanding and are largely limited to evaluating the ability of models to reason about videos, where questions can often be answered without iterative visual interaction. In contrast, \textit{\ourbench} (bottom) introduces questions that require active evidence gathering through a Chain of Manipulations, enabling systematic evaluation of interactive, manipulation-based video reasoning.}
  \label{fig:benchmark}
\end{figure*}

\subsection{Need for \ourbench}
Existing video understanding benchmarks~\cite{li2023videomme, tempcompass, mmvu2024, zhou2025mlvu} are not designed for interactive video reasoning. As shown in Figure~\ref{fig:benchmark}, their main emphasis is on holistic, global video understanding and they do not require the model to iteratively gather and validate visual evidence during the reasoning process. These datasets typically evaluate how well models can reason about videos once visual features are encoded, without requiring iterative perceptual interaction. Consequently, models can often answer questions by relying on static context. In contrast, \textit{\ourbench} introduces questions that necessitate thinking with videos, actively gathering and refining evidence through a Chain of Manipulations, making it possible to assess how well models can engage in interactive reasoning.

\subsection{Benchmark Statistics}
The benchmark comprises $1,000$ high-quality questions derived from $282$ unique videos spanning a diverse range of real-world scenes, activities, and durations. The videos vary from short to long formats, with the majority falling within the mid-range. Specifically, $19$\% of the videos are short ($2$-$4$ minutes), $51$\% are medium-length ($4$–$8$ minutes), and the remaining 30\% are long videos ($8$–$30$ minutes). This variation directly supports manipulation-specific reasoning. Short videos often require precise spatial operations such as frame localization or zooming into transient details, while longer sequences encourage multi-step reasoning across temporally dispersed events, where the model must revisit or compare distant moments.

Each benchmark sample includes a video, a question, five multiple-choice options, the correct answer, and a structured reasoning trace following the exploratory-reasoning → visual-manipulation → observation structure. Each example is annotated with all valid temporal segments, valid frames, and spatial localization tubelets capturing the full temporal span where the target object appears. All annotations and reasoning traces are human-verified to ensure accuracy and consistency.
\section{Qualitative Analysis}

\subsection{GRPO vs RA-GRPO}
\begin{figure*}[!t]
\vspace{-0.5em}
  \centering
  \includegraphics[width=\textwidth]{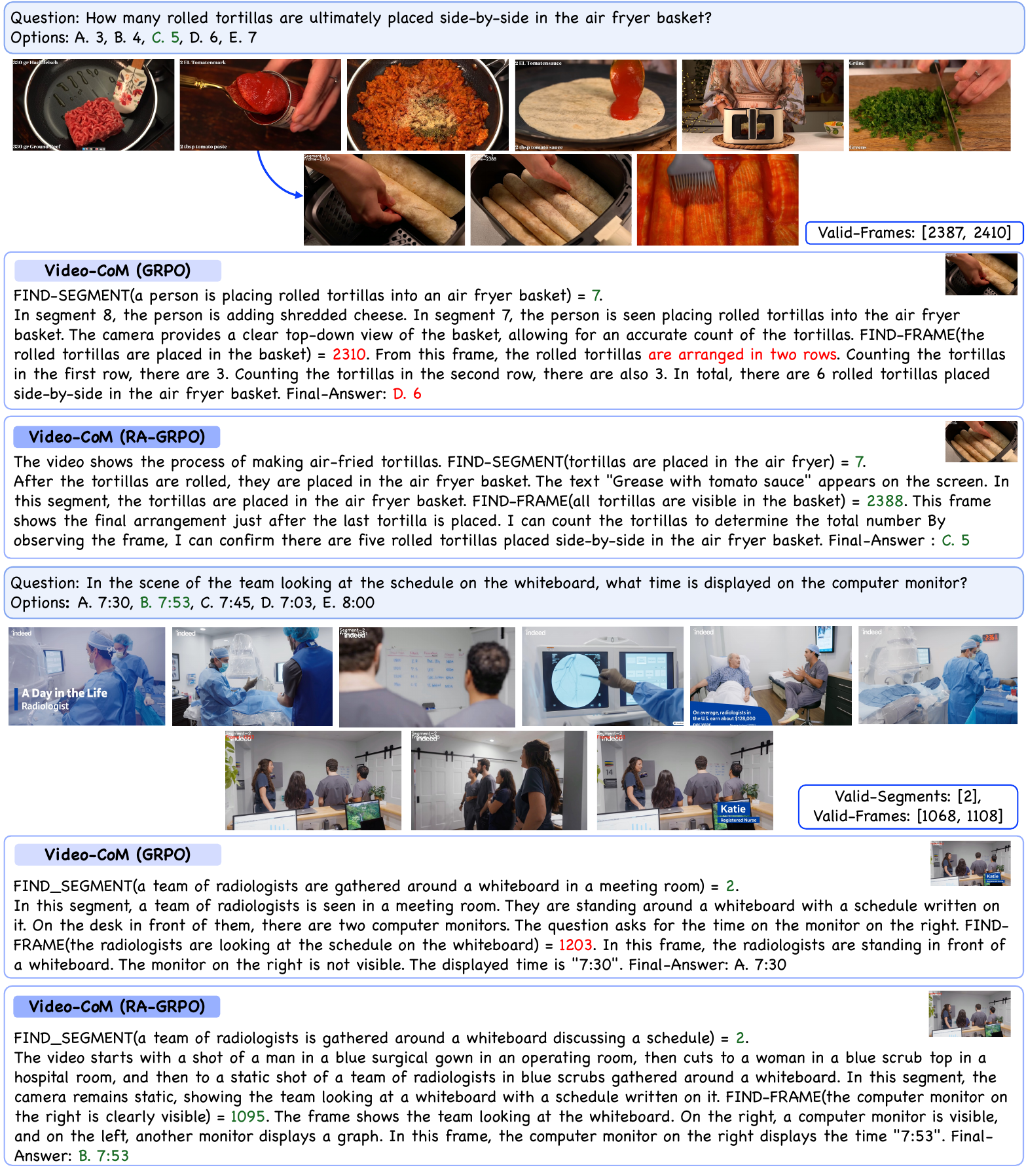}
  \caption{\textbf{Qualitative Comparison between \textit{\ourmodel} with accuracy-only GRPO and reasoning-aware RA-GRPO.}
The examples illustrates how rewarding only the final answer (GRPO) often leads to incomplete or weakly grounded manipulations, resulting in incorrect predictions. In contrast, RA-GRPO explicitly evaluates intermediate manipulations. As shown in the examples (monitor showing the correct time), small variations in frame selection can significantly affect correctness, and step-level reasoning rewards better guide the model toward more accurate reasoning trajectories.}
  \label{fig:acc_vs_ragrpo}
\end{figure*}

Building on the quantitative analysis in Table~\ref{tab:ablations_ra_grpo}, we further provide qualitative comparisons in Figure~\ref{fig:acc_vs_ragrpo}. GRPO with accuracy-only rewards often yields weakly grounded manipulations. In contrast, RA-GRPO introduces step-level supervision that refines temporal and spatial focus across reasoning steps (See Figure~\ref{fig:acc_vs_ragrpo}).

\subsection{Analysis of Visual Attention} 
\begin{figure*}[!t]
  \centering
  \includegraphics[width=\textwidth]{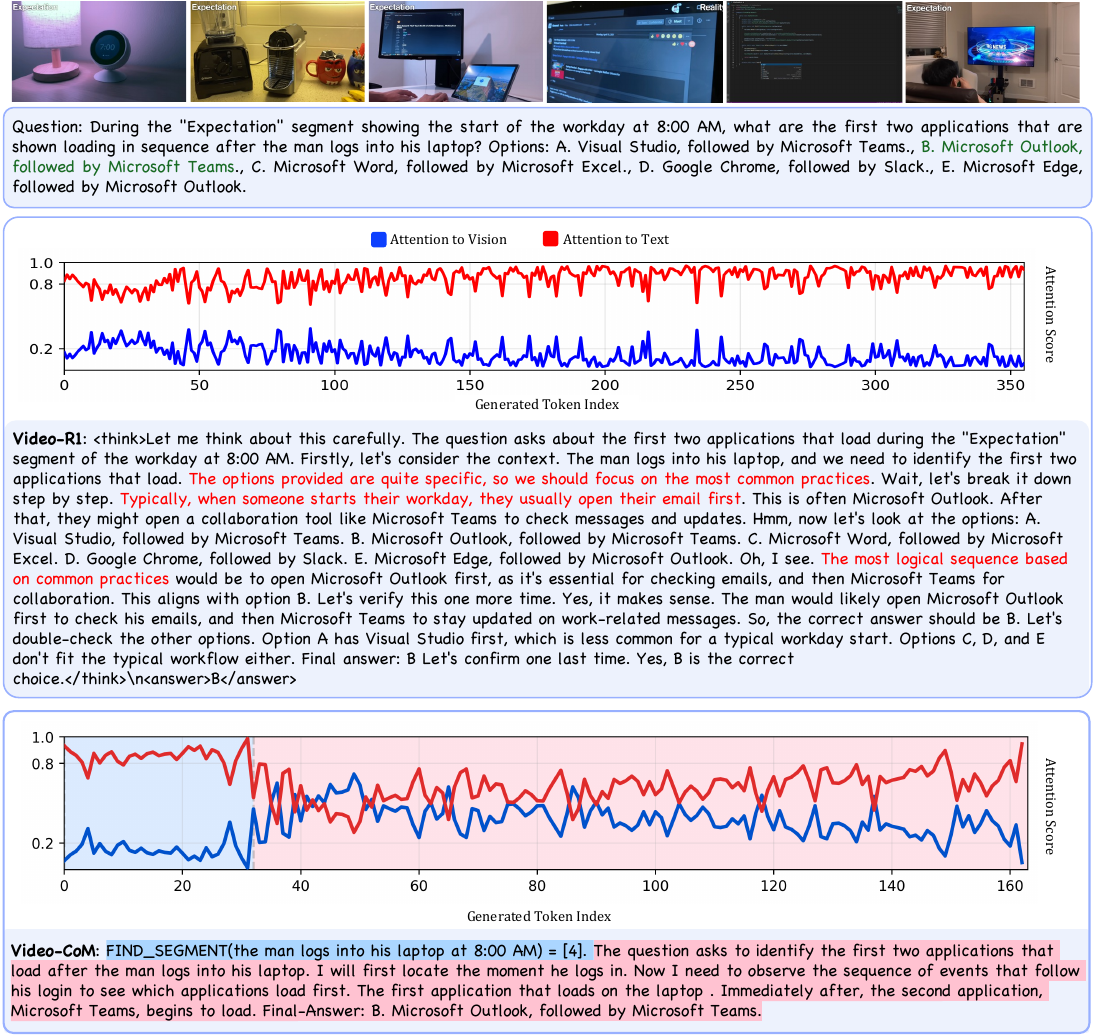}
  \caption{\textbf{Attention Behavior during Visual Reasoning (Example 1)} Video-R1 \cite{feng2025videor1} (top) attends primarily to text tokens, relying heavily on world knowledge about common work routines instead of engaging with the actual visual cues in the video. This behavior shows a strong dependence on textual priors, leading the model to infer answers from general expectations rather than evidence from the scene. Video-CoM (bottom) maintains focused visual attention within the \textit{Find-Segment} frames, aligning with the login and application-loading sequence where the two applications appear in order. The temporally zoomed segment lets the model concentrate on the most relevant portion of the video, capturing subtle visual transitions that support accurate reasoning grounded in visual evidence.}
  \label{fig:visual_attn_1}
\end{figure*}

\begin{figure*}[!t]
  \centering
  \includegraphics[width=\textwidth]{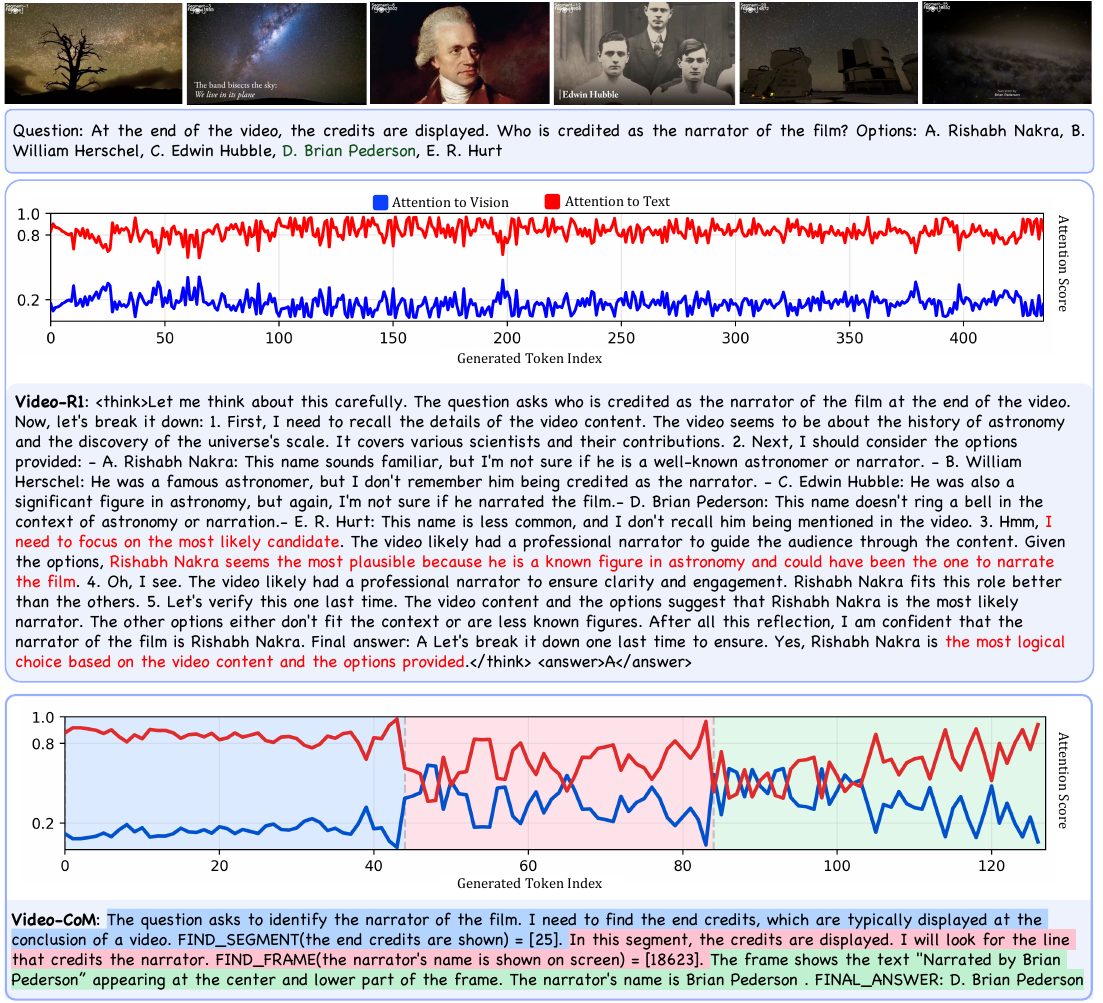}
  \caption{\textbf{Attention Behavior during Visual Reasoning (Example 2)}
Video-R1 \cite{feng2025videor1} (top) gives limited attention to the visual input, forming a biased interpretation that “the video seems to be about the history of astronomy and the discovery of the universe’s scale.” This context shapes its reasoning, leading it to rely on world knowledge and associative cues rather than looking at the relevant visual tokens. The resulting answer is a guess drawn from familiarity instead of direct observation. Video-CoM (bottom) maintains targeted visual attention within the FIND\_SEGMENT frames and applies FIND\_FRAME to pinpoint a clear frame showing “Narrated by Brian Pederson.” By focusing on the temporal and spatial regions relevant to the question, the model verifies the answer visually and avoids drifting into context-based inference.}
  \label{fig:visual_attn_3}
\end{figure*}

We provide additional examples in Figure~\ref{fig:visual_attn_1} and Figure~\ref{fig:visual_attn_3} for analyzing how \textit{Video-CoM} grounds its reasoning in visual evidence. The model consistently maintains high attention to vision tokens during manipulation steps, reflecting active engagement with the visual content. These examples further demonstrate that \textit{Video-CoM} preserves localized visual grounding throughout reasoning, reinforcing its ability to dynamically align perception with evolving reasoning trajectories.

\clearpage
{
    \small
    \bibliographystyle{ieeenat_fullname}
    \bibliography{main}
}

\end{document}